\documentclass{article}

\PassOptionsToPackage{numbers}{natbib}

\usepackage{float}

\usepackage[final]{neurips_2019}

\usepackage[utf8]{inputenc} 
\usepackage[T1]{fontenc}    
\usepackage{hyperref}       
\usepackage{url}            
\usepackage{booktabs}       
\usepackage{amsfonts}       
\usepackage{nicefrac}       
\usepackage{microtype}      
\usepackage{graphicx}
\usepackage[toc,page]{appendix}
\usepackage{caption}
\usepackage{amsmath}

\makeatletter
\newcommand{\customlabel}[2]{%
   \protected@write \@auxout {}{\string \newlabel {#1}{{#2}{\thepage}{#2}{#1}{}} }%
   \hypertarget{#1}{#2}
}
\makeatother

\DeclareMathOperator*{\Argmax}{argmax}
\def\transpose#1{{\mathbf{\mbox{${#1}^T$}}}}
\def\transposed#1{{\mathbf{\mbox{${#1}^\dagger$}}}}
\def\vek#1{{\boldsymbol{#1}}}

\newcommand{\dsp}{\displaystyle}
\newcommand{\Tr}{\mbox{Tr}}
\newcommand{\vst}{\par\vspace{2.5mm}}
\newcommand{\be}{\begin{equation}}
\newcommand{\ee}{\end{equation}}
\newcommand{\bea}{\begin{eqnarray}}
\newcommand{\eea}{\end{eqnarray}}

\newcommand{\beNN}{\begin{equation*}}
\newcommand{\eeNN}{\end{equation*}}
\newcommand{\beaNN}{\begin{eqnarray*}}
\newcommand{\eeaNN}{\end{eqnarray*}}

\newcommand{\W}{{\mathbf{W}^{21}}}
\newcommand{\w}{{\mathbf{W}^{32}}}
\newcommand{\wW}{{\mathbf{W}}}
\newcommand{\Wbar}{{\mathbf{\overline{W}}^{21}}}
\newcommand{\wbar}{{\mathbf{\overline{W}}^{32}}}
\newcommand{\wWbar}{{\mathbf{\overline{W}}}}
\newcommand{\wWbase}{\widetilde{\wW}}
\newcommand{\rank}{\mbox{{\footnotesize rank}}}

\newcommand{\wWleft}{\wW_{>}}
\newcommand{\wWright}{\wW_{<}}

\newcommand{\wA}{{{\mathbf{W}_A}^{\hspace*{-2mm}32}}}
\newcommand{\wB}{{{\mathbf{W}_B}^{\hspace*{-2mm}32}}}

\newcommand{\noisyTeacher}{{\boldsymbol{\hat{\Sigma}}}}
\newcommand{\noisyV}{{\boldsymbol{\hat{V}}}}
\newcommand{\noisyVt}{\transpose{\boldsymbol{\hat{V}}}}
\newcommand{\noisyVtt}{\transposed{\boldsymbol{\hat{V}}}}
\newcommand{\noisyU}{{\boldsymbol{\hat{U}}}}
\newcommand{\noisyS}{{\boldsymbol{\hat{S}}}}
\newcommand{\noisys}{{\boldsymbol{\hat{s}}}}

\newcommand{\SwW}{{\mathbf{S}}}
\newcommand{\SwWbar}{{\mathbf{\overline{S}}}}

\newcommand{\sbarA}{\overline{s}^{A}}

\newcommand{\U}{{\mathbf{\overline{U}}}}
\newcommand{\V}{{\mathbf{\overline{V}}}}
\newcommand{\Vt}{\transpose{\V}}

\newcommand{\bU}{\boldsymbol{U}}

\newcommand{\bS}{\boldsymbol{S}}
\newcommand{\bs}{\boldsymbol{s}}

\newcommand{\Nclasses}{N_{\mathrm{classes}}}
\newcommand{\Nf}{N_{\mathit{f}}}

\newcommand{\ssW}{\boldsymbol{s}^{21}}

\def\ssw#1{{\boldsymbol{s}^{32}_{#1}}}

\newcommand{\x}{{\mathbf{x}}}
\newcommand{\X}{{\mathbf{X}}}

\newcommand{\ybar}{{\mathbf{\overline{y}}}}
\newcommand{\yhat}{{\boldsymbol{\hat{y}}}}
\newcommand{\Y}{{\mathbf{Y}}}

\newcommand{\noise}{\boldsymbol{\xi}}
\newcommand{\Xv}{\vec{X}}
\newcommand{\covX}{\mathbf{C}_{X}}
\newcommand{\crosscov}{\mathbf{C}_{YX}}
\newcommand{\R}{\mathbb{R}}
\newcommand{\Ltrain}{\mathcal{L}_\mathrm{train}}

\newcommand{\order}{\mathcal{O}}

\newcommand{\N}{\overline{N}}

\newcommand{\G}{\mathbf{G}}

\newcommand{\Ndata}{{N_{\mathrm{data}}}}

\newcommand{\argmax}{\mathrm{argmax}}
\def\argmax#1{\mathrm{argmax}_{\mathrm{rows of \mbox{$#1$}}}}

\def\argmax#1{\mbox{${\displaystyle
{\Argmax_{\mathrm{over\,rows}}}\,\big\lbrace #1 \big\rbrace}$}}
\newcommand{\softmax}{\mathrm{softmax}}
\newcommand{\benefit}{\mbox{$MT_{A\leftarrow B}$}}
\newcommand{\relatedness}{\mbox{$r_{AB}$}}
\newcommand{\relatednessMatrix}{\boldsymbol{r}}

\newlength{\mylen}
\setbox1=\hbox{$\bullet$}\setbox2=\hbox{\tiny$\bullet$}
\title{Generalization in multitask deep neural classifiers: a statistical physics approach}

\author{
 Tyler Lee \\
  Intel AI Lab\\
  \texttt{tyler.p.lee@intel.com} \\
  \And
  Anthony Ndirango \\
  Intel AI Lab \\
  \texttt{anthony.ndirango@intel.com} \\
}

\begin{document}

\maketitle

\begin{abstract}
A proper understanding of the striking generalization abilities of deep neural networks presents an enduring puzzle. Recently, there has been a growing body of numerically-grounded theoretical work that has contributed important insights to the theory of learning in deep neural nets. There has also been a recent interest in extending these analyses to understanding how multitask learning can further improve the generalization capacity of deep neural nets. These studies deal almost exclusively with regression tasks which are amenable to existing analytical techniques. We develop an analytic theory of the nonlinear dynamics of generalization of deep neural networks trained to solve classification tasks using softmax outputs and cross-entropy loss, addressing both single task and multitask settings. We do so by adapting techniques from the statistical physics of disordered systems, accounting for both finite size datasets and correlated outputs induced by the training dynamics. We discuss the validity of our theoretical results in comparison to a comprehensive suite of numerical experiments. Our analysis provides theoretical support for the intuition that the performance of multitask learning is determined by the noisiness of the tasks and how well their input features align with each other. Highly related, clean tasks benefit each other, whereas unrelated, clean tasks can be detrimental to individual task performance.
\end{abstract}

\section{Introduction}
Despite the remarkable string of successful results demonstrated by deep learning practitioners, we still do not have a clear understanding of how these models manage to generalize so well, effectively evading many of the intuitions expected from statistical learning theory. The enigma is further heightened when one considers multitask learning, especially in regimes where labeled data is scarce. In order to make specific assertions about the effective transfer of knowledge across tasks, one needs a predictive framework to address generalization in a multitask setting. There has been a noticeable uptick in recent efforts to build a rigorous theoretical foundation for deep learning (see, e.g. \cite{AdvaniSaxe2017, Saxe_etal2013, Mallat2016, LinTegmark2016, Salmhofer_etal2018, SagunBiroli2018, ChayesSagunZecchina,  GoldtSaxe2019, Lee_etal2019, Arora_etal2019} for a sampling of this trend). To the best of our knowledge (with one exception, described below), none of the existing analytical work deals with multitask learning. 

Multitask learning holds promise for training more generalized and intelligent learning systems \cite{Caruana1997}. It comprises a broad set of strategies loosely defined by the presence of multiple objective functions and a set of shared parameters optimized for those objective functions. The most prevalent formulation of multitask learning in the literature is the addition of supervised auxiliary task(s) to assist in training a network to better perform a target task of interest (\textit{main task})\cite{Qian2015, Luong2015, Kim2017,Liu2019}. In this framework the only purpose of the auxiliary task(s) is to produce improved generalization performance on the main task. This benefit is thought to arise from an inductive bias placed on the learning of the main task towards learning more general features \cite{Caruana1997}. Since the features learned through multitask learning blend the optimal features for all of the optimized tasks, there is an assumed dependence of the multitask benefit on the relatedness of the auxiliary tasks to the main task (e.g. if the optimal features for the auxiliary task are orthogonal to those of the main task, then the main task will be best optimized by ignoring the auxiliary task entirely). How exactly to define "relatedness" in the context of multitask learning in deep neural networks remains unknown. The most explicit definition to date, to our knowledge, comes from \cite{LampinenGanguli2019}, where it is described as the angles between the singular vectors of the implicit input-output function learned by the network. While this definition is narrow, it lends a nice starting point for a theoretical analysis in the multitask setting. Outside of the work done in \cite{LampinenGanguli2019} on multitask learning in linear regression networks, the theory of multitask learning in neural networks remains unexplored. In this work we hope to further the theoretical understanding of multitask benefits to multiclass classification problems, a much more common class of problems in modern machine learning.

To narrow the scope of this study, we have chosen to focus on the formulation of multitask learning where the neural network is defined as having a single shared trunk and multiple task-specific heads. Many recent studies have sought to explore alternative methods of parameter sharing, though these do not usually lend themselves as easily to this form of theoretical analysis \cite{MisraSGH16,MeyersonM2017}. Further, multitask learning also provides an interesting strategy for learning a single universal representation for many tasks possibly across multiple domains \cite{BilenV17,Howard2018,KaiserGSVPJU17}. In this strategy there is often no clear "main" task and it is not clear that the benefit to be gained is even improved generalization performance on any of the trained tasks. Instead the benefit could be seen as improved performance over a set of problems given a fixed parameter budget or improved transfer learning to unseen tasks \cite{DoerschandZisserman2017}. While these are certainly exciting research directions and could benefit from careful theoretical scrutiny, we leave them for future work.

This manuscript is structured as follows: in section \ref{single_theory} we describe the theory behind single task learning in classification networks. In section \ref{single_training} we describe, both analytically and empirically, the training dynamics of such networks. In section \ref{multi_teacher} we extend this work to account for multitask learning of simple classification tasks. Finally, in section \ref{discussion} discuss interesting leads and future directions.

\section{Theoretical Underpinnings}\label{single_theory}
A convenient framework for analyzing multitask problems was introduced in  \cite{LampinenGanguli2019}, addressing regression problems in deep linear neural networks. Given the success of that approach, could the techniques  in \cite{LampinenGanguli2019} be generalized to deep neural net classifiers with softmax outputs?  Our analysis provides an affirmative answer to this question, albeit at considerable technical cost: despite a strong conceptual similarity between analyzing regression and softmax classification problems, the structure of the solutions to the classification problem differ markedly from those obtained in the regression case. On the other hand, and perhaps unsurprisingly, the intuition gleaned from  \cite{LampinenGanguli2019} about the conditions required for effective multitask learning carry over to the classification problems, in spite of the technical differences between the analysis of classification and regression tasks.

We adopt the student-teacher setup popularized several decades ago in early attempts to theoretically understand the generalization abilities of neural networks (see, e.g. \cite{BosKinzelOpper1993}) and recently revisited in \cite{LampinenGanguli2019}. We will attempt to closely follow the notational conventions in \cite{LampinenGanguli2019} with the hope of establishing a common language for analyzing these sorts of problems. 
The key insight behind the analysis of softmax classifiers is the uncanny resemblance of the training dynamics of deep neural nets to the physical dynamics of disordered systems. In particular, we take advantage of a formal similarity between deep neural softmax classifers and a generalized version of Derrida's Random Energy Model (REM) \cite{Derrida_REM}. A generalization of the REM is required because the outputs of a deep neural network are correlated random variables, in contrast to the i.i.d conditions that render the original REM solvable. Furthermore, deep learning practitioners do not work with infinite size models, so we also have to take into account finite size effects. 

\subsection{Teacher Network}
Following \cite{LampinenGanguli2019}, we consider low rank teacher networks which serve to provide a training signal to arbitrary student networks. We begin with a 3-layer teacher network defined by 
$\overline{N}_\ell$ units in layer $\ell$ and weight matrices $\Wbar \in \mathbb{R}^{\N_2\times\N_1}$ between the input and hidden layer and $\wbar \in \mathbb{R}^{\N_3\times\N_2}$ between the hidden layer and an {\em argmax} output layer. We also define $\wWbar \equiv \wbar\Wbar \in \mathbb{R}^{\N_3\times\N_1}$ for the teacher's composite weight.

We consider teachers that produce noisy outputs using a noise perturbed composite weight matrix $\noisyTeacher \equiv \wWbar+ \noise$, where $\noise \in \mathbb{R}^{\N_3\times\N_1}$ has i.i.d elements.

During training, the teacher network takes in an input data matrix $\X\in\mathbb{R}^{\N_1\times\Ndata}$, and produces noisy vector outputs $\displaystyle{\yhat \equiv \argmax{\noisyTeacher\X} \in \mathbb{R}^{\Ndata}}$

thereby furnishing a rule for producing (noisy) labels $\yhat$ from inputs $\X$. At test time, the student is tested against noise-free labels generated via 
$\displaystyle{
\ybar \equiv \argmax{\wWbar\X} \in \mathbb{R}^{\Ndata}}$

At this point, we take a slight departure from the setup in \cite{LampinenGanguli2019}: in their setup, the data matrix is taken to be orthonormal, whereas we take $\X$ to have entries 
drawn independently  from a standard Gaussian distribution. Similarly, the elements of the noise matrix $\noise$ are i.i.d centered normal variables with variance $\hat{\sigma}^2/\N_1$. The scale of 
$\hat{\sigma}$ is chosen in such a way that there is a non-zero probability for label-flipping, i.e. 
$\mathrm{Prob}(\yhat \neq \ybar)>0$. 

\subsection{Student Network}
We first consider a 3-layer student network. In general, the student network has the same number of input and output units as the teacher since these are defined by the specifics of the task at hand. However, the student has no knowledge of the teacher's internal architecture. Thus, the number of hidden units in the student's network will almost surely be different from the teacher's. Writing $N_2$ for the student's number of hidden units, we have student weight matrices $\W \in \mathbb{R}^{N_2\times\N_1}$ between the input and hidden layer and $\w \in \mathbb{R}^{\N_3\times N_2}$ between the hidden layer and the softmax output layer. We also define $\wW \equiv \w\W \in \mathbb{R}^{\N_3\times\N_1}$ for the student's composite weight.

Given an input data matrix $\X\in\mathbb{R}^{\N_1\times\Ndata}$,  the student computes a matrix output 
$$\Y(\wW\X)=\softmax\big(\wW\X\big)$$ 
Note that  $\Y\in\mathbb{R}^{\N_3\times \N_1}$ is a matrix with elements

$$
\Y_{c\mu}(\wW\X) = \softmax\left(\sum_{k=1}^{\N_1}\wW_{ck}\X_{k\mu}\right), \qquad 1 \le c \le \N_3, \, 1 \le \mu \le \Ndata
$$

which is interpreted as the probability that the student assigns a class label $c$ given an input $\x^{\mu}$ drawn from the $\mu$$^{\mathrm{th}}$ column of $\X$.

The student is trained by minimizing a cross-entropy loss

\begin{equation}
\mathcal{L}_\mathrm{train}
=-
\frac{1}{\Ndata}\sum_{\mu=1}^{\Ndata}\sum_{c=1}^{\N_3}\delta_{c, \yhat_{\mu}(\X)}\ln{\Y_{c\mu}(\wW\X)}, \qquad \mbox{(where $\delta$ is the Kronecker delta.)}
\label{train_loss}
\end{equation}

%

\section{Training Dynamics: Theory v/s Experiment}\label{single_training}
We use vanilla SGD to train the student network. A detailed derivation of the dynamics of training is presented in appendix~\ref{appendix:dynamics}. The relevant equations are given by

\begin{eqnarray}
\nonumber
\tau \frac{d}{dt}\w &=& \Big( \G(\noisyTeacher) \noisyTeacher - \G(\wW)\wW \Big) \transpose{\W} \\
\tau \frac{d}{dt}\W &=&  \transpose{\w}\Big( \G(\noisyTeacher) \noisyTeacher - \G(\wW)\wW \Big)
\label{SGD}
\end{eqnarray}

where $1/\tau$ is the SGD learning rate, and $\G: \mathbb{R}^{\N_3\times\N_1} \mapsto \mathbb{R}^{\N_3\times\N_3}$ is a non-linear, {\em positive semi-definite} matrix-valued function which captures the gradient of the softmax function averaged over the training data (see appendix~\ref{appendix:dynamics}:\ref{Gmatrix_defn} for a precise definition). 
The solutions to (\ref{SGD}) are very different from those obtained for the regression case in \cite{LampinenGanguli2019}. 

Further insight into the dynamics (\ref{SGD}) is provided by considering the so-called {\em training aligned} (TA) case as defined in \cite{LampinenGanguli2019} where one initializes the student's weights such that 
the initial value of the student's composite weight is $\wW_0 = \noisyU\SwW_0\noisyVt$ given the noisy teacher's SVD $\noisyTeacher =   \noisyU \,\noisyS\, \noisyVt$, where $\SwW_0$ is the student's initial singular value matrix. 

A detailed analysis of the TA dynamics is presented in full generality in appendix~\ref{appendix:TAmodel}. For a rank one teacher in the TA case, i.e. if the noisy teacher's SVD is $\noisyTeacher=\hat{s}\vek{\hat{u}}\transpose{\vek{\hat{v}}}$, equation (\ref{SGD}) simplifies further to an equation for the student's largest singular value, with all the other singular values exponentially suppressed in time. Explicitly, writing $s \equiv \max{\SwW}$ for the student's largest singular value,  equation (\ref{SGD}) becomes

\begin{equation}
\tau \frac{d}{dt}s = 2s\vek{\hat{u}}\cdot\Big( \hat{s}\G(\hat{s}\vek{\hat{u}}\transpose{\vek{\hat{v}}}) 
- s\G(s\vek{\hat{u}}\transpose{\vek{\hat{v}}}) \Big)  \vek{\hat{u}}
\label{SGDTA}
\end{equation}

\begin{figure}
\centering
\includegraphics[]{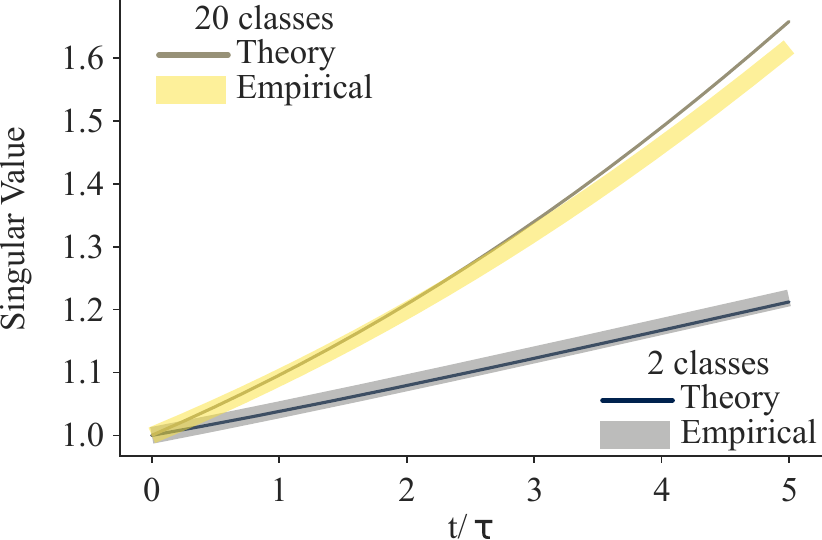}
\caption{
Comparing the theoretical predictions in (\ref{SGDTA}) to empirical results. $1/\tau=10^{-3}$ is the learning rate, so the figure shows training for 5k steps (chosen as the minimum of the validation error). 
The empirical results are obtained using 10 different random seeds. The results shown are for a 2-class and 20-class classification task using 100 training data points to highlight the fact that the theory agrees with experiment over a wide range of class sizes.
}
\label{TA_dynamics}
\end{figure}

Numerically integrating equation (\ref{SGDTA}) yields the graphs shown in Figure~\ref{TA_dynamics}. The figure reveals excellent agreement between theory and experiment over a wide range of initial conditions.

\section{Multitask Generalization Dynamics: Theory v/s Experiment}\label{multi_teacher}
\subsection{Teacher Networks}
In the multitask setting, we have two teacher networks represented by $\N_3\times \N_1$ weight matrices  $\wWbar_A$ and $\wWbar_B$ with ranks 
$\N_2^{A}$ and $\N_2^{B}$ respectively. Their noise-perturbed versions, $\noisyTeacher_A, \, \noisyTeacher_B$ are defined as before, so that the teachers produce noisy labels 
$\yhat_{A/B} \equiv \argmax{\noisyTeacher_{A/B}\X}$ and noise free labels $\ybar_{A/B} \equiv \argmax{\wWbar_{A/B}\X}$. 

\subsection{Student Network}
In the multitask setting, a composite student network is designed to learn multiple tasks jointly from the teachers. In general, the student network will consist of a trunk comprised of a stack of hidden layers shared across tasks, augmented by a set of specialized heads specific to individual tasks. This setup is identical to the one used in \cite{LampinenGanguli2019}.

For three-layer students, we continue to denote the trunk's composite weight matrix by $\W$ and write 
$\wA$, $\wB$ for the weights in the heads, and $\wW_A \equiv \wA\W$,   $\wW_B \equiv \wB\W$ for the corresponding composite weights. Note that, crucially, both students share the trunk weights $\W$.

The students are trained to minimize a weighted sum of the cross-entropy losses pertaining to each task, i.e. 
$\mathcal{L} = \alpha_A \mathcal{L}_A +  \alpha_B \mathcal{L}_B$. In general, the weighting coefficients $\alpha_A,\, \alpha_B$ can be chosen via some optimization method or even learned as part of the model's training procedure. However, we will only consider the simplest case where $\alpha_A=\alpha_B=1$.

We arbitrarily pick task A as the main task that we're interested in, and consider task B as an auxiliary task whose sole purpose is to improve the performance of task A. We are thus interested in finding out what properties of task B are required in order to improve the student's learning of task A. This naturally leads to the idea of {\em task-relatedness}, a well-known, though loosely-defined, concept in the literature on multitask learning \cite{Caruana1997}. 

\subsection{Task Relatedness}
As noted in the introduction, we currently lack a precise definition of task-relatedness in the context of multitask learning in deep neural networks. The authors of \cite{LampinenGanguli2019} propose defining task-relatedness as a function of the angles between the singular vectors of the implicit input-output function learned by the network. As it turns out, as a direct consequence of the SGD dynamics in~(\ref{SGD}),  the same definition appears naturally in the student-teacher framework for multitask classifiers. 

Given two tasks $A$ and $B$ defined by two teachers with weight matrices $\wWbar_A$ and $\wWbar_B$ respectively, we denote their SVDs by 
$\displaystyle{\wWbar_{A/B} = \U_{A/B}\,\SwWbar_{A/B}\,\Vt\hspace{-3mm}\,_{A/B.}}$ We define the relatedness $\relatednessMatrix_{AB}$ between tasks $A$ and $B$ as 

\be
\relatednessMatrix_{AB} := \Vt\hspace{-2mm}_B\overline{\boldsymbol{V}}_A
\ee

\subsection{Multitask Benefit}
\begin{table}[h!]
    \centering
    \caption{Key takeaways from multitask analysis}
    \label{tbl:takeaways}
    \resizebox{\textwidth}{!}{      
    \begin{tabular}{ c c c c c p{5cm}}
     & \multicolumn{3}{ c }{independent variables}  & &\\ \cline{2-4}
         & $r_{AB}$ & $\overline{s}_B $ & $\Ndata$  & effect on $\benefit$ & analytical explanation\\ 
        \toprule
        \customlabel{exp:unrelated}{(a)} & 0 & any & any & 0 & $s_A =  \widetilde{s}_A$\\ 
        \customlabel{exp:clean_aux}{(b)} & > 0 & $\nearrow$ & any & $\nearrow$
            & $(s_A - \widetilde{s}_A) \searrow$ as $\overline{s}_B \nearrow$\\   
        \customlabel{exp:medium_main}{(c)} & $r_{AB} \nearrow \mbox{\small{$(0 < r_{AB} \ll 1)$}}$ & any & limited & $\nearrow$  & {\small appendix:\ref{appendix:TArank1benefit}, eqn.~(\ref{benefit_vs_rAB})} \\   
        \customlabel{exp:large_main}{(d)} & any & any & abundant & small & $\widetilde{s}_A g(\widetilde{s}_A)   \rightarrow \overline{s}_Ag(\overline{s}_A)$\\
        \bottomrule
    \end{tabular}
   }
\end{table}


For the purposes of quantifying any gains in performance from multitask learning relative to models trained on a single task, we introduce the notion of a {\em multitask benefit}.  We arrive at our multitask benefit by comparing the optimal performance of the multitask model on the main task, say $A$ to the optimal performance of a {\em baseline} model trained only on task $A$.

Given the multitask generalization loss $\mathcal{L}_{AB} = \mathcal{L}_A +  \mathcal{L}_B$, we define  $\mathcal{L}_{A\vert B} := \mathcal{L}_{AB} - \mathcal{L}_B$ as the generalization loss on task $A$ when task $A$ is trained jointly with task B. This quantity is to be compared to the generalization loss $\widetilde{\mathcal{L}}_A$ defined as the loss when task A is trained on its own. Following \cite{LampinenGanguli2019}, we define the multitask benefit conferred on task A by task B via 

$$
\benefit \equiv \min_{t} \left\lbrace  \widetilde{\mathcal{L}}_A(t) \right\rbrace -
\min_{t} \left\lbrace \mathcal{L}_{A\vert B}(t)\right\rbrace
$$

Remarkably, one can place a tight bound on the multitask benefit using a relatively simple argument based on the concavity of the logarithm function. 
We present here the result for the simpler case of a TA model with rank one teachers and relegate the general case to appendix~\ref{appendix:mtlbenefit}. 
For a TA model with rank one teachers with SVD 
$\wWbar_A=\overline{s} \vek{u}_{A} \transpose{\vek{v}_{A}}$, we abbreviate $g(s) :=  \vek{u}_{A} \cdot \G(s\vek{u}_{A}\transpose{\vek{v}_{A}})  \vek{u}_{A} \ge 0$, 
with $\G$ as featured in the training dynamics in equation (\ref{SGD}) and defined in appendix~\ref{appendix:dynamics}:\ref{Gmatrix_defn}. The key takeaways of this analysis are summarized in Table \ref{tbl:takeaways} and described more fully below.

As derived in Appendix~\ref{appendix:mtlbenefit} (cf. equations C:\ref{benefit_lower} and C:\ref{benefit_upper}), the bound on the multitask benefit is 

\be
\left(s_A - \widetilde{s}_A\right) 
\Big( 
\overline{s}_A g(\overline{s}_A)  - s_A g(s_A)\Big) 
\,\, \le \benefit \le  \,\,
\left(s_A - \widetilde{s}_A\right) 
\Big( 
\overline{s}_A g(\overline{s}_A)  - \widetilde{s}_A g(\widetilde{s}_A)\Big) 
\label{rank1benefit}
\ee

\vst
Notice that the factor $\big( \overline{s}_A g(\overline{s}_A)  - \widetilde{s}_A g(\widetilde{s}_A)\big)$
on the RHS of equation (\ref{rank1benefit}) depends only quantities pertaining to the baseline single task case, and hence is entirely independent of the training dynamics of the multitask case.  

In contrast, the sign of $\left(s_A - \widetilde{s}_A\right)$ depends on the multitask teachers' singular values for tasks A and B, their correspponding SNRs, and the relatedness $\relatedness$ between tasks A and B (see the discussion surrounding equations~\ref{singularvalsdynamics}-\ref{infinite_data} in Appendix~\ref{appendix:TArank1benefit}). 
For unrelated tasks, {\em viz.} $\relatedness = 0$, one obtains $s_A = \widetilde{s}_A$ (cf.~\ref{appendix:TArank1benefit}:\ref{singularvalsdynamics}) and so the multitask benefit vanishes. For ``weakly related'' tasks, {\em viz.} $0 < \relatedness \ll 1$,~(\ref{appendix:TArank1benefit}:\ref{benefit_vs_very_large_sB}) shows that high SNR auxiliary tasks have a deleterious effect on $\benefit$.  

In the high SNR regime, the noisy teacher's singular values are larger than the noise-free case. Since the student's dynamics is driven by the noisy teacher, $s_A \to \hat{s}_A \ge \overline{s}_A$ in the high SNR regime. Under these conditions, equation~(\ref{appendix:TArank1benefit}:\ref{hiSNR}) implies that 
$\benefit \ge 0$.

In the low SNR regime, the noisy teacher's singular values lie in the bulk of the MP sea \cite{BenaychGeorges}. In this case, the student's dynamics is driven by noise, so that $s_A \to \hat{s}_A < \overline{s}_A$ for low SNRs. Under these conditions, a positive $\benefit$ occurs only if the constraints on $\relatedness$ and $\overline{s}_B$ leading to equation~(\ref{appendix:TArank1benefit}:\ref{benefit_lowSNR}) are satisfied. 

In regimes where labeled training data is abundant, the factor $\big( \overline{s}_A g(\overline{s}_A)  - \widetilde{s}_A g(\widetilde{s}_A)\big) \rightarrow 0$ in which case $\benefit \rightarrow 0$, regardless of the relatedness between tasks (cf.~equation~\ref{appendix:TArank1benefit}:\ref{infinite_data}). 

To summarize, the TA model predicts that multitask learning will have the largest impact under conditions mimicking scarce labeled data such that the baseline model underperforms on the main task, as long as the auxiliary tasks have some relatedness to the main task. Thus, coming up with auxiliary tasks that have a high degree of relatedness to the main task will be crucial to observing a positive multitask benefit. 

While the results in this section have only been demonstrated for the special case of TA models, we will shortly see that the predictions are realized empirically in a wide variety of scenarios.

\subsection{Data vs model uncertainty}
\begin{figure}
  \centering
  \includegraphics[]{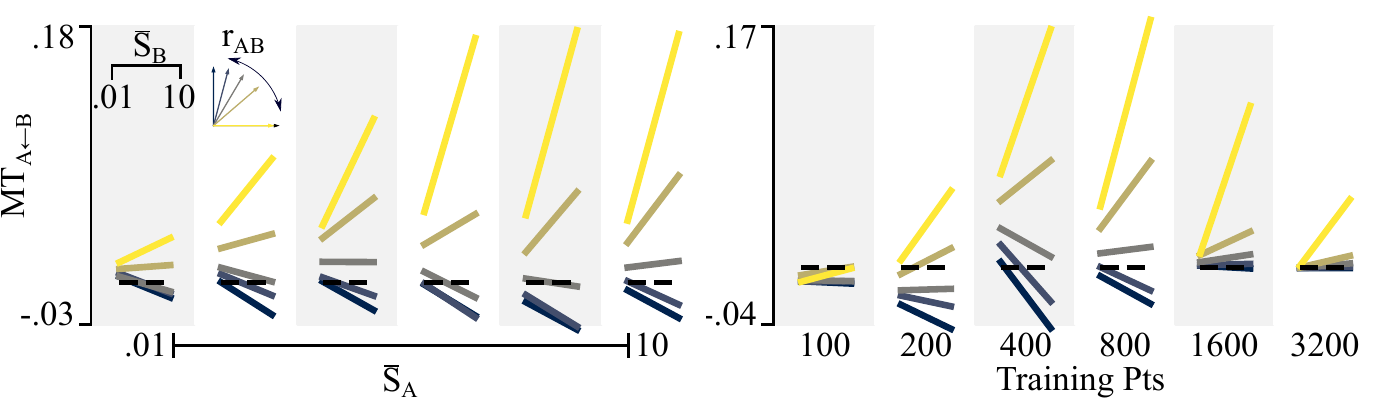}
  \caption{(Left) Summary of multitask benefits gained when the student network was trained with increasing signal-to-noise ratio (SNR). With constant noise levels, the SNR increases with the singular values for teacher A, $\overline{\boldsymbol{S}}_A$, were increased from .01 to 10 (alternating stripes, left-to-right). For each value of $\overline{\boldsymbol{S}}_A$ (x-axis), the average multitask benefit was computed for low SNR auxiliary tasks ($\overline{\boldsymbol{S}}_B$) and high SNR auxiliary tasks (each line segment, left-to-right) across 5 levels of task relatedness ($r_{AB}$). Data is plotted for 800 training points.  This demonstrates that multitask benefit is correlated with task relatedness and SNR for related tasks, yet negatively correlated with SNR for unrelated tasks. (Right) Summary of multitask benefits with increasing amount of training data (alternating stripes, left-to-right). At 100 training points the network still struggles to train and does not gain a generalization benefit from auxiliary data. For > 200 training points, the network begins to leverage the related auxiliary data to improve performance. When the dataset is very large, performance nearly reaches its ceiling and the auxiliary data has little effect. See Figure \ref{appendix:rank10} for the complete set of interactions among these variables.
  }
  \label{fig:multi_main}
\end{figure}

Using the framework described above, we set out to describe the relationship between multitask benefit and several key factors that influence training of both the single task baseline - the amount and quality of the main task data - and multitask training - the amount, quality and relatedness of auxiliary task data. We systematically varied\footnote{Code supporting this paper is available upon request} these factors and computed the multitask benefit for 5 different training datasets, the results of which are summarized in Figure~\ref{fig:multi_main}. To ensure that we had roughly class-balanced training datasets, we fixed $\N_3 = \N_2$, and set both to 10 for the experiments here. Other values for the rank showed similar results and data for rank 3 teacher networks can be found in Figure~\ref{appendix:rank3}.  The signal-to-noise ratio (SNR) of the data in each dataset is directly proportional to the singular value of the teacher network that generated each task's data.

 We kept all singular values for a given teacher network the same and varied this value from .01 to 100. Similarly, we fixed the relatedness of teacher network B to $\Vt\hspace{-2mm}_B\overline{\boldsymbol{V}}_A = \relatedness I$, such that the singular vectors $\overline{\boldsymbol{V}}_B$ were orthogonal to $\overline{\boldsymbol{V}}_A$ with constant inner product. We varied this value from 0 to 1. This work demonstrates several interesting dependencies: 
 
\begin{enumerate}
    \item  Multitask benefit increases with increasing task relatedness and SNR of the auxiliary data. This mirrors the finding from row \ref{exp:clean_aux} of Table \ref{tbl:takeaways}.
    \item Unrelated, high SNR auxiliary tasks are actually destructive to the learning process of the main task. Our theoretical framework provides an explanation for this observation in~\ref{appendix:TArank1benefit}:\ref{benefit_vs_very_large_sB}. In contrast, unrelated, noisy auxiliary tasks are readily ignored. This mirrors the findings from rows \ref{exp:unrelated} and \ref{exp:medium_main} of Table \ref{tbl:takeaways}.
    \item The main task must have a certain level of base performance either from clean data or larger amounts of data before multitask learning can help. This holds up to the point where single task performance nears optimal performance on the main task, as is the case when the amount of training data supplied is large. These statements mirror the findings from rows \ref{exp:medium_main} and \ref{exp:large_main} of Table \ref{tbl:takeaways}.
\end{enumerate}

\begin{figure}[t]
  \centering
  \includegraphics[]{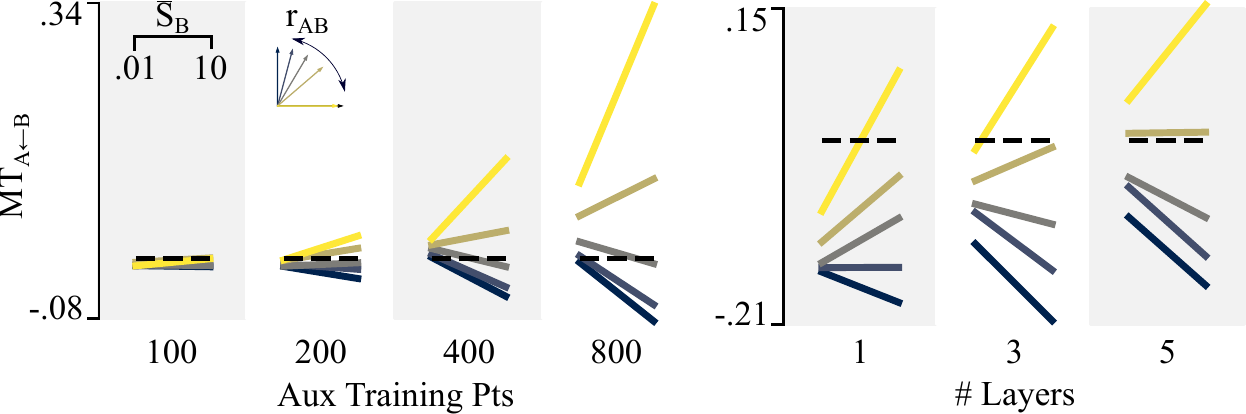}
  \caption{(Left) Summary of multitask benefits gained when the student network was trained with increasing amounts of auxiliary task data . For each quantity of auxiliary task data (x-axis), the average multitask benefit was computed for low SNR aux tasks and high SNR aux tasks (each line segment, left-to-right) across 5 levels of task relatedness. All the data shown is for high SNR main tasks, and demonstrates that increasing relatedness and auxiliary task data give large multitask benefits. For more details see Figure~\ref{appendix:multi_aux_data}. (Right) Summary of multitask benefits gained for nonlinear student networks of increasing depth (x-axis). Deeper nonlinear networks show similar trends to shallow linear networks. For more details see Figure~\ref{appendix:multi_nonlinear}.}
  \label{fig:multi_aux_data}
\end{figure}
\subsection{Auxiliary task data efficiency}
Multitask learning is a popular strategy for extending the utility of a limited amount of main task data. This is often an interesting choice when auxiliary task data is easy to come by but main task data is expensive. To gauge the value of additional auxiliary task data while holding main task data fixed, we trained multitask student networks on 100 main task data points and up to 800 auxiliary task data points. These results are summarized in Figure~\ref{fig:multi_aux_data} (left) and full results can be found in  Figure~\ref{appendix:multi_aux_data}. As auxiliary task data quantities increase we see similar trade-offs to those above, where related, high quality data provides a large multitask benefit, while unrelated, high quality data proves increasingly detrimental.

\subsection{Multitask learning in deeper, nonlinear student networks}
To ensure that our results can generalize to nonlinear and deeper networks, we varied the number of hidden layers in the student network and included a ReLU nonlinearity between each hidden layer. While this situation does not lend itself to clean theoretical analysis, we found that these networks behave qualitatively similar to the linear network results described above. These results are summarized in Figure~\ref{fig:multi_aux_data} (right) and full results can be found in Figure~\ref{appendix:multi_nonlinear}. Again, multitask benefit is strongly correlated with relatedness and the SNR of both datasets. Interestingly, there is a general shift downwards in multitask benefit, suggesting that nonlinear networks require more highly related tasks in order to generate a significant performance increase.

\section{Discussion and future directions}\label{discussion}
Here we demonstrate that, for linear classifier networks with a softmax output nonlinearity, generalization performance can be computed analytically. We extend the analysis in \cite{LampinenGanguli2019} to classification problems and show both theoretically and empirically that improvements from multitask learning are heavily related to training set size, task relatedness, and the noise levels inherent in the data. Networks given sufficient data to train well show improved performance when supplemented with related, high signal-to-noise ratio auxiliary tasks. Unrelated auxiliary tasks show little benefit and can be actively detrimental if they provide a strong enough training signal. 

The problem of increasing the range of parameters from which one gets a multitask benefit and decreasing potential harms has received increasing interest in recent years, often through clever loss or gradient weighting strategies \cite{KendallGC17,Sener2018,Du2018}. A careful interrogation of ~(\ref{rank1benefit}) should provide some insight on methods for maximizing the possible multitask benefit, a direction we leave for future work. Additionally, we have shown that our results generalize to deeper, more nonlinear student networks, though these networks are still quite different from networks used in practice. We expect the insights gained in this work, especially with regard to the critical properties of main and auxiliary task datasets will generalize well to more complex networks. Generalizing our results regarding task relatedness poses an interesting challenge for future research.

\subsubsection*{Acknowledgments}
We would like to thank Cory Stephenson, Gokce Keskin, Oguz Elibol, Suchismita Padhy, and Ting Gong for many fruitful discussions regarding this work. We must also acknowledge Nicholas Sapp for his work in establishing the compute infrastructure that made the empirical portions of this work possible.

\bibliography{neurips_mtl_2019.bib}{}

\begin{thebibliography}{10}

\bibitem{AdvaniSaxe2017}
Madhu~S. Advani and Andrew~M. Saxe.
\newblock High-dimensional dynamics of generalization error in neural networks.
\newblock {\em CoRR}, abs/1710.03667, 2017.

\bibitem{Saxe_etal2013}
Andrew~M. Saxe, James~L. McClelland, and Surya Ganguli.
\newblock Exact solutions to the nonlinear dynamics of learning in deep linear
  neural networks.
\newblock In {\em 2nd International Conference on Learning Representations,
  {ICLR} 2014, Banff, AB, Canada, April 14-16, 2014, Conference Track
  Proceedings}, 2014.

\bibitem{Mallat2016}
St{\'{e}}phane Mallat.
\newblock Understanding deep convolutional networks.
\newblock {\em CoRR}, abs/1601.04920, 2016.

\bibitem{LinTegmark2016}
Henry~W. Lin and Max Tegmark.
\newblock Why does deep and cheap learning work so well?
\newblock {\em CoRR}, abs/1608.08225, 2016.

\bibitem{Salmhofer_etal2018}
Felix Dr{\"{a}}xler, Kambis Veschgini, Manfred Salmhofer, and Fred~A.
  Hamprecht.
\newblock Essentially no barriers in neural network energy landscape.
\newblock In {\em Proceedings of the 35th International Conference on Machine
  Learning, {ICML} 2018, Stockholmsm{\"{a}}ssan, Stockholm, Sweden, July 10-15,
  2018}, pages 1308--1317, 2018.

\bibitem{SagunBiroli2018}
Marco Baity{-}Jesi, Levent Sagun, Mario Geiger, Stefano Spigler,
  G{\'{e}}rard~Ben Arous, Chiara Cammarota, Yann LeCun, Matthieu Wyart, and
  Giulio Biroli.
\newblock Comparing dynamics: Deep neural networks versus glassy systems.
\newblock In {\em Proceedings of the 35th International Conference on Machine
  Learning, {ICML} 2018, Stockholmsm{\"{a}}ssan, Stockholm, Sweden, July 10-15,
  2018}, pages 324--333, 2018.

\bibitem{ChayesSagunZecchina}
Pratik Chaudhari, Anna Choromanska, Stefano Soatto, Yann LeCun, Carlo Baldassi,
  Christian Borgs, Jennifer~T. Chayes, Levent Sagun, and Riccardo Zecchina.
\newblock Entropy-sgd: Biasing gradient descent into wide valleys.
\newblock In {\em 5th International Conference on Learning Representations,
  {ICLR} 2017, Toulon, France, April 24-26, 2017, Conference Track
  Proceedings}, 2017.

\bibitem{GoldtSaxe2019}
Sebastian Goldt, Madhu~S. Advani, Andrew~M. Saxe, Florent Krzakala, and Lenka
  Zdeborov{\'{a}}.
\newblock Generalisation dynamics of online learning in over-parameterised
  neural networks.
\newblock {\em CoRR}, abs/1901.09085, 2019.

\bibitem{Lee_etal2019}
Jaehoon Lee, Lechao Xiao, Samuel~S. Schoenholz, Yasaman Bahri, Jascha
  Sohl{-}Dickstein, and Jeffrey Pennington.
\newblock Wide neural networks of any depth evolve as linear models under
  gradient descent.
\newblock {\em CoRR}, abs/1902.06720, 2019.

\bibitem{Arora_etal2019}
Sanjeev Arora, Simon~S. Du, Wei Hu, Zhiyuan Li, Ruslan Salakhutdinov, and
  Ruosong Wang.
\newblock On exact computation with an infinitely wide neural net.
\newblock {\em CoRR}, abs/1904.11955, 2019.

\bibitem{Caruana1997}
Rich Caruana.
\newblock Multitask learning.
\newblock {\em Machine Learning}, 28(1):41--75, Jul 1997.

\bibitem{Qian2015}
Y.~Qian, M.~Yin, Y.~You, and K.~Yu.
\newblock Multi-task joint-learning of deep neural networks for robust speech
  recognition.
\newblock In {\em 2015 IEEE Workshop on Automatic Speech Recognition and
  Understanding (ASRU)}, pages 310--316, Dec 2015.

\bibitem{Luong2015}
Minh-Thang {Luong}, Quoc~V. {Le}, Ilya {Sutskever}, Oriol {Vinyals}, and Lukasz
  {Kaiser}.
\newblock {Multi-task Sequence to Sequence Learning}.
\newblock {\em arXiv e-prints}, page arXiv:1511.06114, Nov 2015.

\bibitem{Kim2017}
Suyoun Kim, Takaaki Hori, and Shinji Watanabe.
\newblock {Joint CTC-attention based end-to-end speech recognition using
  multi-task learning}.
\newblock {\em ICASSP, IEEE International Conference on Acoustics, Speech and
  Signal Processing - Proceedings}, pages 4835--4839, 2017.

\bibitem{Liu2019}
Xiaodong Liu, Pengcheng He, Weizhu Chen, and Jianfeng Gao.
\newblock Multi-task deep neural networks for natural language understanding.
\newblock {\em CoRR}, abs/1901.11504, 2019.

\bibitem{LampinenGanguli2019}
Andrew~Kyle Lampinen and Surya Ganguli.
\newblock An analytic theory of generalization dynamics and transfer learning
  in deep linear networks.
\newblock {\em CoRR}, abs/1809.10374, 2018.

\bibitem{MisraSGH16}
Ishan Misra, Abhinav Shrivastava, Abhinav Gupta, and Martial Hebert.
\newblock Cross-stitch networks for multi-task learning.
\newblock {\em CoRR}, abs/1604.03539, 2016.

\bibitem{MeyersonM2017}
Elliot Meyerson and Risto Miikkulainen.
\newblock Beyond shared hierarchies: Deep multitask learning through soft layer
  ordering.
\newblock {\em CoRR}, abs/1711.00108, 2017.

\bibitem{BilenV17}
Hakan Bilen and Andrea Vedaldi.
\newblock Universal representations: The missing link between faces, text,
  planktons, and cat breeds.
\newblock {\em CoRR}, abs/1701.07275, 2017.

\bibitem{Howard2018}
Jeremy Howard and Sebastian Ruder.
\newblock Fine-tuned language models for text classification.
\newblock {\em CoRR}, abs/1801.06146, 2018.

\bibitem{KaiserGSVPJU17}
Lukasz Kaiser, Aidan~N. Gomez, Noam Shazeer, Ashish Vaswani, Niki Parmar, Llion
  Jones, and Jakob Uszkoreit.
\newblock One model to learn them all.
\newblock {\em CoRR}, abs/1706.05137, 2017.

\bibitem{DoerschandZisserman2017}
Carl Doersch and Andrew Zisserman.
\newblock Multi-task self-supervised visual learning.
\newblock {\em CoRR}, abs/1708.07860, 2017.

\bibitem{BosKinzelOpper1993}
S.~B\"os, W.~Kinzel, and M.~Opper.
\newblock Generalization ability of perceptrons with continuous outputs.
\newblock {\em Phys. Rev. E}, 47:1384--1391, Feb 1993.

\bibitem{Derrida_REM}
Bernard Derrida.
\newblock Random-energy model: An exactly solvable model of disordered systems.
\newblock {\em Phys. Rev. B}, 24:2613--2626, Sep 1981.

\bibitem{BenaychGeorges}
Florent Benaych-Georges and Raj~Rao Nadakuditi.
\newblock The singular values and vectors of low rank perturbations of large
  rectangular random matrices.
\newblock {\em Journal of Multivariate Analysis}, 111:120--135, 2012.

\bibitem{KendallGC17}
Alex Kendall, Yarin Gal, and Roberto Cipolla.
\newblock Multi-task learning using uncertainty to weigh losses for scene
  geometry and semantics.
\newblock {\em CoRR}, abs/1705.07115, 2017.

\bibitem{Sener2018}
Ozan Sener and Vladlen Koltun.
\newblock Multi-task learning as multi-objective optimization.
\newblock {\em CoRR}, abs/1810.04650, 2018.

\bibitem{Du2018}
Yunshu {Du}, Wojciech~M. {Czarnecki}, Siddhant~M. {Jayakumar}, Razvan
  {Pascanu}, and Balaji {Lakshminarayanan}.
\newblock {Adapting Auxiliary Losses Using Gradient Similarity}.
\newblock {\em arXiv e-prints}, page arXiv:1812.02224, Dec 2018.

\bibitem{HLP}
J.E.~Littlewood G.H.~Hardy and G.P{\`o}lya.
\newblock {\em Inequalities}.
\newblock Cambridge University Press UK, 1934.

\end{thebibliography}
\bibliographystyle{unsrt}
\clearpage
\begin{appendices}

\renewcommand{\thefigure}{A\arabic{figure}}
\setcounter{figure}{0}

\section*{Notation}
\begin{itemize}
\item
Given a matrix $\mathbf{A}$, we will denote its transpose by $\mathbf{A}^{\dagger}$.
\item
Given a pair of random vectors $\vec{X}$, $\vec{Y}$, we will denote their cross covariance matrix by $\crosscov$.
\item
Given a pair of vectors $\vec{u}\in\R^m$, $\vec{v} \in\R^n$, we define 
$\vec{u}\otimes\vec{v} \in \R^{m \times n}$ as the matrix with entries $(\vec{u}\otimes\vec{v})_{ak} := u_a v_k$.
\item
Given a pair $\vec{u}$, $\vec{v}$ of $n$-dimensional vectors  we denote their Hadamard product by 
$\vec{u}\odot\vec{v} \in \R^{n}$, i.e. $(\vec{u}\odot\vec{v})_{k} := u_k v_k$. 
\item
${O}(N):=$  group of $N\times N$ orthogonal matrices.
\end{itemize}
\section{Teacher-Student Setup}
\label{appendix:dynamics}

\subsection{Teacher Network}
We consider teachers defined by a weight matrix  $\wWbar \in \mathbb{R}^{\Nclasses\times\Nf}$, where $\Nclasses$ is the number of classes and $\Nf$ the number of input features. 
Noisy teachers are defined by a weight matrix $\noisyTeacher \equiv \wWbar+ \noise$, where $\noise \in  \mathbb{R}^{\Nclasses\times\Nf}$ has entries drawn independently from a 
centered Gaussian distribution with variance $\hat{\sigma}^2/\Nf$.

During training, the teacher network takes in an input data matrix $\X\in\mathbb{R}^{\Nf\times\Ndata}$, and produces noisy vector outputs 
$$
\displaystyle{\yhat \equiv \argmax{\noisyTeacher\X} \in \mathbb{R}^{\Ndata}}
$$

thereby furnishing a rule for producing (noisy) labels $\yhat$ from inputs $\X$. At test time, the student is tested against noise-free labels generated via 
$\displaystyle{
\ybar \equiv \argmax{\wWbar\X} \in \mathbb{R}^{\Ndata}}$.

The columns of  $\X$ form a collection of $\Ndata$ feature vectors $\lbrace \vec{X}^{\mu} \rbrace$, $\mu=1, \cdots, \Ndata$, drawn from a centered Gaussian distribution with covariance $\covX$. We will write $\hat{y}(\vec{X}^{\mu})$ for the label assigned to the feature vector $\vec{X}^{\mu}$. We assume that the matrix $\X$ is of full rank so that $\X^\dagger\X$ is invertible.

\subsection{Student Network}
A student network with $L$ layers is defined via a collection of weight matrices $\wW^{(l)} \in \mathbb{R}^{N_{l}\times N_{l-1}}$, $1\le l\le L$, with $N_0=\Nf$ and $N_L=\Nclasses$. 
The student's composite weight matrix is given by 
$
\wW := \wW^{(L)}\wW^{(L-1)} \cdots \wW^{(1)}.
$

Define 
\beaNN
\wWleft^{(l)} &:=& \wW^{(L)} \cdots \wW^{(l+1)},\\
\wWright^{(l)} &:=& \wW^{(l-1)}\cdots \wW^{(1)}
\eeaNN

so that, for $2\le l < L$, 
$$
\wW = \wWleft^{(l)}\wW^{(l)}\wWright^{(l)}.
$$

In particular, the gradient of any scalar valued function $f(\wW)$,  with respect to $\wW^{(l)}$ is given by

\be
\nabla_{\wW^{(l)}} f = 
 {\wWleft^{(l)}}^{\dagger} \left(\nabla_\wW f\right) {\wWright^{(l)}}^{\dagger}
\label{gradf}
\ee

Let $\mathbb{P}_c(\wW\vec{X})$ define the probability of observing class $c$ given $\wW\vec{X}$. For a neural classifier, this reads

$$
\mathbb{P}_c(\wW\vec{X}) := \mathrm{Probability}\left(\mathrm{class} \,c\, \big\vert \wW \vec{X} \right) := \softmax{[\wW\vec{X}][c]}.
$$

The cross-entropy loss between the teacher's one-hot-distributed labels $\lbrace \hat{y}(\vec{X}^{\mu})) \rbrace$ and the student's softmax outputs can be written as 

\beaNN
\Ltrain(\wW|\noisyTeacher, \X)
&=&-
\frac{1}{\Ndata}\sum_{\mu=1}^{\Ndata}
\ln{ \mathbb{P}_{\hat{y}(\vec{X}^{\mu})}(\wW\vec{X}^{\mu})}
\\ 
&=& - \frac{1}{\Ndata}\sum_{\mu=1}^{\Ndata}\sum_{c=1}^{\Nclasses}
\mathbb{P}_c(\beta\noisyTeacher\vec{X}^{\mu})
\ln{ \mathbb{P}_c(\wW\vec{X}^{\mu})}
\eeaNN

where $\beta\gg1$ is a parameter chosen such that $\mathbb{P}_c(\beta\noisyTeacher\vec{X}^{\mu})$ is arbitrarily close to the noisy teacher's outputs $\hat{y}(\vec{X}^{\mu})$.

\subsection{Training Dynamics}
The student's weights are updated layerwise via SGD. Adopting the ``continuous time'' version of SGD for ease of exposition, and using the identity (\ref{gradf}),  the layerwise update equations read 

\beaNN
\tau\frac{d}{dt}\wW^{(l)}
&=& -\nabla_{\wW^{(l)}} \Ltrain(\wW|\noisyTeacher, \X) \\
&=&  
- {\wWleft^{(l)}}^{\dagger}
 \left[
\frac{1}{\Ndata}
\sum_{\mu=1}^{\Ndata}
\sum_{c=1}^{\Nclasses}
\mathbb{P}_c(\beta\noisyTeacher\vec{X}^{\mu})
 \left(\nabla_\wW \ln{ \mathbb{P}_c(\wW\vec{X}^{\mu} )} \right) 
 \right]
 {\wWright^{(l)}}^{\dagger}
\eeaNN

A straightforward calculation reveals that 

\beaNN
\frac{1}{\Ndata}
\sum_{\mu=1}^{\Ndata}
\sum_{c=1}^{\Nclasses}
\mathbb{P}_c(\beta\noisyTeacher\vec{X}^{\mu})
\left(\nabla_\wW \ln{ \mathbb{P}_c(\wW\vec{X}^{\mu} )} \right) 
&=&
\crosscov(\wW) - 
\crosscov(\beta\noisyTeacher)
\eeaNN

where the matrix $\crosscov(\wW)$ defined by

\be
\crosscov(\wW)_{c,k} = 
\frac{1}{\Ndata}
\sum_{\mu=1}^{\Ndata}
\mathbb{P}_c(\wW\vec{X}^{\mu})X^{\mu}_k
\label{crosscov}
\ee

is the student's estimate of the empirical cross-covariance between the softmax outputs and the feature vectors determined using the training dataset. Similarly, $\crosscov(\beta\noisyTeacher)$ is the 
empirical cross-covariance between the feature vectors and the labels generated by the teacher. 

Therefore, 

\be
\tau\frac{d}{dt}\wW^{(l)}
=
{\wWleft^{(l)}}^{\dagger} 
\Big[  
\crosscov(\beta\noisyTeacher) - \crosscov(\wW)
\Big]
{\wWright^{(l)}}^{\dagger}
\label{layerWupdate}
\ee

Equation (\ref{layerWupdate}) yields an interesting relationship between the weights in consecutive layers, {\em viz}.

\be
\frac{d}{dt}\left[{\wW^{(l+1)}}^\dagger\wW^{(l+1)}\right] 
= \frac{d}{dt}\left[\wW^{(l)}{\wW^{(l)}}^\dagger\right]
\qquad 1 \le l \le L-1
\label{pairs}
\ee

Using equation (\ref{layerWupdate}), a straightforward calculation gives

\be
\tau\frac{d}{dt} \Ltrain(\wW|\noisyTeacher, \X)
=
-\sum_{l=1}^{L}
\Tr
\left(
{\wWright^{(l)}}^{\dagger}{\wWright^{(l)}}
\left[\crosscov(\beta\noisyTeacher) - \crosscov(\wW)\right]^{\dagger}
{\wWleft^{(l)}}{\wWleft^{(l)}}^{\dagger} 
\left[\crosscov(\beta\noisyTeacher) - \crosscov(\wW)\right]
\right)
\label{dLdt}
\ee

Each summand on the RHS of equation (\ref{dLdt}) is the trace of a product of symmetric positive semi-definite matrices. Hence $\dsp{\frac{d\mathcal{L}_\mathrm{train}}{dt} \le 0}$ throughout training. Thus, SGD is guaranteed to converge to a solution which minimizes  $\Ltrain( \cdot |\noisyTeacher, \X)$, although we have not provided any information about the rate of convergence.

Furthermore, 

$$
\min_{\wW}{\Ltrain(\wW|\noisyTeacher, \X)} = \Ltrain(\noisyTeacher|\noisyTeacher, \X).
$$

and 

$$
\mbox{$\wW$ is a minimum of $\Ltrain(\,\cdot\,|\noisyTeacher, \X)$} \,\,\, \Leftrightarrow \,\,\, \crosscov(\wW) = \crosscov(\beta\noisyTeacher). 
$$

In other words, the optimal solutions include all cases where the student's estimate of the empirical cross-covariance matches that of the noisy teacher. The number of such solutions is highly degenerate due to the fact that the softmax function is invariant under all transformations $\wW \to \wW + \vec{1}\otimes\vec{v}$ for any vector $\vec{v}\in \mathbb{R}^{\Nf}$, where $\vec{1}\in\mathbb{R}^{\Nclasses}$ is the vector of all ones.

A straightforward computation shows that the Hessian of  $\Ltrain(\,\cdot\,|\noisyTeacher, \X)$ has only non-negative eigenvalues, which combined with equation (\ref{dLdt}) leads to the conclusion that the set of minima of the loss is given by

\be
\left\lbrace \wW = \beta\noisyTeacher +  \lambda\sum_{\mu=1}^{\Ndata}\vec{1}\otimes(\X^\dagger\X)^{-1}\vec{X}^{\mu} \,\,\,\forall \lambda \in \R \right\rbrace.
\label{optima}
\ee

\subsubsection{Training in the limit of infinite data}
Finally, we note that as $\Ndata \to \infty$, equation (\ref{crosscov}) reads 

\be
\lim_{\Ndata\to\infty}
\frac{1}{\Ndata}
\sum_{\mu=1}^{\Ndata}
\mathbb{P}_c(\wW\vec{X}^{\mu})X^{\mu}_k
\to 
\mathbb{E}\left(X_k\mathbb{P}_c(\wW\vec{X})\right)
\label{crosscov2}
\ee 
where $ \mathbb{E}(\cdot)$ denotes the expectation over $\vec{X}$. When $\vec{X}$ is a centered Gaussian random vector with covariance $\covX$, then Gaussian integration by parts in (\ref{crosscov2}) yields

$$
\mathbb{E}\left(X_k\mathbb{P}_c(\wW\vec{X})\right) = [\G(\wW)\wW\covX]_{ck}
$$

where the matrix $\G(\wW)$ is defined as 

\bea
\G(\wW)_{c, c^{\prime}} := \mathbb{E} \left(\mathbb{P}_c(\wW\vec{X})\right)\delta_{c, c^{\prime}} -  \mathbb{E} \left(\mathbb{P}_c(\wW\vec{X})\mathbb{P}_{c^{\prime}}(\wW\vec{X}) \right).
\label{Gmatrix_defn}
\eea

Thus, from equations (\ref{crosscov})  and (\ref{crosscov2}), we obtain 

\be
\lim_{\Ndata\to\infty} \crosscov(\wW) =  \G(\wW)\wW\covX.
\label{crosscovToG}
\ee

We note that, from the definition in (\ref{Gmatrix_defn}), $\G(\wW)$ is a positive semi-definite matrix with a single zero eigenvalue. Furthermore, if the diagonal entries of $\wW\wW^\dagger$ are much larger in magnitude than its off diagonal entries, then one can combine the HLP theorem~\cite{HLP} with a generalization of Derrida's REM techniques~\cite{Derrida_REM}~to show that:

\begin{enumerate}
\item

$$
\G(\wW) \simeq \frac{1}{\Nclasses-1}\Tr(\G(\wW)) \left(\mathbf{I} - \frac{1}{\Nclasses}\vec{1}\otimes\vec{1}\right).
$$

\item
If the SVD of $\wW$ is given by $\wW=\mathbf{USV^{\dagger}}$, then 
$\Tr(\G(\wW)) \simeq g(\mathbf{S})$ where the explicit functional form of the real-valued function $g$ can be accurately estimated for large values of the norm $\Vert \mathbf{S} \Vert$ of $\mathbf{S}$. Under the stated conditions, one can show that 

\be
\mbox{the individual components, $g(\bS)S_{\alpha\alpha}$, decrease monotonically with the norm $\Vert \bS \Vert$.}
\label{dgdS}
\ee

\item
Furthermore, 
\be
\mathbf{U}^\dagger
\G(\mathbf{USV^{\dagger}})\mathbf{U} \simeq \frac{1}{\Nclasses-1}  g(\mathbf{S}) \mathbf{I}.
\label{approxG}
\ee

\end{enumerate}

Surprisingly, our empirical results obtained over a wide range of experimental conditions suggest that using the above approximate equalities gives very accurate results even in regimes where we include the off-diagonal entries of $\wW\wW^\dagger$. 
In other words, the corrections obtained by including the off-diagonal entries are always marginal in our experiments.

\section{Training Aligned (TA) Networks}
\label{appendix:TAmodel}
We now specialize the results in the previous section to the so-called TA networks \cite{LampinenGanguli2019}\footnote{Our definition of TA networks differs slightly from the TA networks in \cite{LampinenGanguli2019}.}. TA networks are a class of analytically tractable models where one can explicitly calculate the quantities appearing in equations (\ref{layerWupdate}, \ref{dLdt}, \ref{Gmatrix_defn}, and \ref{approxG}). 

The key point is that TA networks are defined only by the choice of initialization of model parameters, and we are free to choose the initial values of these parameters to make the model solvable. Of course, in reality, deep learning practitioners do not have access to an oracle as in the student-teacher setup, so any initialization that 
assumes knowledge of the teachers' SVD is not feasible in practice. Nevertheless, simulations show that the intuition gained from TA models generalizes to networks initialized randomly. 

For our TA model, we assume that we are using an SVD convention where the $\noisyU$ and $\noisyV$ are orthogonal matrices and the singular value matrix is rectangular with zeros off the main diagonal. Given the teacher's SVD 
$\beta\noisyTeacher = \noisyU \noisyS \noisyVtt$, we choose a set of orthogonal matrices $\lbrace \bU^{(l)} \rbrace_{l=0}^{L}$ with $ \bU^{(l)} \in {O}(N_l)$,  $\bU^{(L)} := \noisyU$, $\bU^{(0)} := \noisyV$, and set 


\beaNN
\wW_0^{(l)} = \bU^{(l)}\bS_0^{(l)}{\bU^{(l-1)}}^{\dagger}, 
\qquad
{\wWright^{(l)}}_0 = \bU^{(l-1)} {\bS_{<}^{(l)}}_0 \noisyVtt, 
\qquad
{\wWleft^{(l)}}_0 = \noisyU {\bS_{>}^{(l)}}_0 {\bU^{(l)}}^{\dagger}
\eeaNN

so that the student's initial composite weight matrix is 

$$
\wW_0 =  {\wWleft^{(l)}}_0 \wW_0^{(l)}{\wWright^{(l)}}_0   = \noisyU  {\bS_{>}^{(l)}}_0  {\bS_{0}^{(l)}}  {\bS_{<}^{(l)}}_0 \noisyVtt
$$

Using the estimate in (\ref{approxG}), the SGD update equations for the TA model at $t=0$ read 

\bea
\left.
\tau\frac{d}{dt} {\bS^{(l)}} \right\vert_{t=0}
&=&
 {\bS_{>}^{(l)}}_0^{\dagger} 
\noisyU^{\dagger} 
\Big[  
\crosscov(\beta\noisyTeacher) - \crosscov(\wW_0)
\Big]
\noisyV {\bS_{<}^{(l)}}_0^{\dagger}
\nonumber\\
&\simeq& 
 {\bS_{>}^{(l)}}_0^{\dagger}
\Big[  
g(\noisyS) \noisyS - g(\bS_0) \bS_0
\Big]
{\bS_{<}^{(l)}}_0^{\dagger}
\label{TAupdate0}
\eea

The RHS of (\ref{TAupdate0}) is a diagonal matrix, so that 

$$
{\bS^{(l)}}(\Delta t) \simeq {\bS_0^{(l)}} + \frac{\Delta t}{\tau}
 {\bS_{>}^{(l)}}_0^{\dagger}
\Big[  
g(\noisyS) \noisyS - g(\bS_0) \bS_0
\Big]
{\bS_{<}^{(l)}}_0^{\dagger} 
+ 
\order{\left(\left( \frac{\Delta t}{\tau}\right)^2\right)}
$$

Thus, repeatedly iterating this construction gives, for arbitrary $t$, 

$$
\frac{d}{dt}\wW^{(l)}
=
 \bU^{(l)}
 \frac{d}{dt} {\bS^{(l)}}
 {\bU^{(l-1)}}^{\dagger} 
$$

with

\bea
\tau\frac{d}{dt} {\bS^{(l)}}
&\simeq& 
 {\bS_{>}^{(l)}}^{\dagger}
\Big[  
g(\noisyS) \noisyS - g(\bS) \bS
\Big]
{\bS_{<}^{(l)}}^{\dagger}
\label{TAupdate1}
\eea

In other words, under the stated assumptions, SGD only modifies the singular values of the weights in each layer, leaving the singular vectors fixed at their initial values. 

We henceforth drop the ``$\simeq$'' and write the equations as equalities. Specializing to the case where $L=2$, equation (\ref{TAupdate1}) becomes

\beaNN
\tau\frac{d}{dt} {\bS^{(2)}}
&=& 
\Big[  
g(\noisyS) \noisyS - g(\bS) \bS
\Big]
{\bS^{(1)}}^{\dagger} 
, \qquad
\tau\frac{d}{dt} {\bS^{(1)}}
=
 {\bS^{(2)}}^{\dagger}
\Big[  
g(\noisyS) \noisyS - g(\bS) \bS
\Big]
\eeaNN

where $\bS = \bS^{(2)}\bS^{(1)}$. If we define $\bs^{(l)}$ as the vector consisting of the non-zero elements of $\bS^{(l)}$, the previous equation reads

\beaNN
\tau\frac{d}{dt} {\bs^{(2)}}
&=& 
\Big[  
g(\noisys) \noisys - g(\bs) \bs
\Big]\odot
{\bs^{(1)}}
, \qquad
\tau\frac{d}{dt} {\bs^{(1)}}
=
\Big[  
g(\noisys) \noisys - g(\bs) \bs
\Big]\odot {\bs^{(2)}}
\eeaNN

where, now $\bs = \bs^{(2)}\odot\bs^{(1)}$. Consequently,

\beaNN
\tau\frac{d}{dt}\bs = 
\tau\frac{d}{dt}\left[ \bs^{(2)} \odot \bs^{(1)}\right]
=
\Big[  
g(\noisys) \noisys - g(\bs) \bs
\Big]
\odot
\Big[
\bs^{(1)}\odot\bs^{(1)} + \bs^{(2)}\odot\bs^{(2)}
\Big]
\eeaNN

Taking into account equation (\ref{pairs}), we have

$$
\bs^{(1)}\odot\bs^{(1)} = \bs^{(2)}\odot\bs^{(2)} + \mbox{constant}
$$

where the constant term is determined by the choice of initial conditions. For simplicity, we pick the initial non-zero singular values to all have the same value so that the constant vanishes. 

$$
\bs^{(1)}\odot\bs^{(1)} = \bs^{(2)}\odot\bs^{(2)} \qquad \Rightarrow \qquad \bs = \bs^{(2)}\odot\bs^{(1)} = \bs^{(1)}\odot\bs^{(1)}. 
$$

Thus, we finally obtain 
\beaNN
\tau\frac{d}{dt}\bs = 
\tau\frac{d}{dt}\left[ \bs^{(2)} \odot \bs^{(1)}\right]
=
2\Big[  
g(\noisys) \noisys - g(\bs) \bs
\Big]
\odot
\bs
\eeaNN

which is equation (\ref{SGDTA}) in our paper.

\section{Multitask Benefit}
\label{appendix:mtlbenefit}

We will derive an expression for the multitask benefit \benefit in the most general setting, assuming Gaussian inputs (not necessarily iid) and linear activations, except for the softmax in the classifier. No other assumptions are required, and the result holds for models of any depth.

We will use the notation $\langle F(\Xv) \rangle$ for  the expectation of $F$ over the distribution of $\vec{X}$, where we assume that  $\vec{X}$ is a centered Gaussian random vector with covariance $\covX$.

The generalization error is obtained by computing 
\bea
\mathcal{L} := \mathcal{L}_\mathrm{generalization}
&=&
-
\left\langle
\sum_{c=1}^{\Nclasses}\delta_{c, \ybar(\vec{X})}
\ln{ \mathbb{P}_c(\wW\vec{X})}
\right\rangle
\\ \nonumber
&=&
-
\sum_{c=1}^{\Nclasses}
\left\langle
\mathbb{P}_c(\beta\wWbar\vec{X})
\ln{ \mathbb{P}_c(\wW\vec{X})} 
\right\rangle
\\ \nonumber
&=&
\langle \ln{Z}(\wW\vec{X})\rangle
-
\sum_{c=1}^{\Nclasses}
\left\langle
\mathbb{P}_c(\beta\wWbar\vec{X})
[\wW\vec{X}]_c 
\right\rangle
\label{test_loss}
\eea

where $Z(\vec{v}):=\dsp{\sum_{c=1}^{\Nclasses} e^{v_c}}$ for any vector $\vec{v}\in\R^{\Nclasses}$. 

Note that we have replaced the noisy teacher's weights $\noisyTeacher$ with the denoised teacher's weights $\wWbar$ since we test the model using the ground truth labels generated from the true distribution.

Using the same notation as in the main paper, we write $\wW_A$ and $\wWbase_A$ for the parameters of task $A$ in the multitask setting and the single task baseline respectively. Thus, the generalization loss on the main task in the 
multitask setting is given by 

$$
\mathcal{L}_{A\vert B} = 
\langle \ln{Z}(\wW_A\vec{X})\rangle
-
\lim_{\beta\to\infty}
\sum_{c=1}^{\Nclasses}
\left\langle
\mathbb{P}_c(\beta\wWbar\vec{X})
[\wW_A\vec{X}]_c 
\right\rangle 
$$

whereas the generalization loss for the baseline model trained on task A is 

\setlength{\belowdisplayskip}{0pt}
$$
\mathcal{L}_{A} = 
\langle \ln{Z}(\wWbase_A\vec{X}) \rangle
-
\lim_{\beta\to\infty}
\sum_{c=1}^{\Nclasses}
\left\langle
\mathbb{P}_c(\beta\wWbar\vec{X})
[\wWbase_A\vec{X}]_c 
\right\rangle 
$$

The multitask benefit is obtained by computing

$$
\benefit := \mathcal{L}_{A} - \mathcal{L}_{A|B} = 
\left\langle\ln\frac{Z(\wWbase_A\vec{X})}{Z(\wW_A\vec{X})} \right\rangle  + 
\lim_{\beta\to\infty}
\sum_{c=1}^{\Nclasses}
\left\langle
\mathbb{P}_c(\beta\wWbar\vec{X})
\Big\lbrace
[\wW_A\vec{X}]_c - [\wWbase_A\vec{X}]_c 
\Big\rbrace
\right\rangle 
$$

Now, $\ln{Z}(\vec{v})$ is convex  in $\vec{v}$ for any vector $\vec{v}$ (since the Hessian of $\ln{Z}$ is a positive definite symmetric matrix). Hence

\bea
\nonumber
\ln\left[\frac{Z(\wWbase_A\vec{X})}{Z(\wW_A\vec{X})} \right] 
&\ge& 
(\wWbase_A - \wW_A)\cdot\nabla_{\wW_A}\ln{Z}(\wW_A) \\ 
&=&
\sum_{c=1}^{\Nclasses}
\mathbb{P}_c(\wW_A\vec{X})
\Big\lbrace
[\wWbase_A\vec{X}]_c - [\wW_A\vec{X}]_c 
\Big\rbrace
\label{lowerbound}
\eea

Interchanging $\wW_A \leftrightarrow \wWbase_A$ yields

\bea
\nonumber
\ln\left[\frac{Z(\wWbase_A\vec{X})}{Z(\wW_A\vec{X})} \right]  
&\le& 
(\wWbase_A - \wW_A)\cdot\nabla_{\wWbase_A}\ln{Z}(\wWbase_A) \\ 
&=&
\sum_{c=1}^{\Nclasses}
\mathbb{P}_c(\wWbase_A\vec{X})
\Big\lbrace
[\wWbase_A\vec{X}]_c - [\wW_A\vec{X}]_c 
\Big\rbrace
\label{upperbound}
\eea

So, taking expectations in (\ref{lowerbound}), we get 

\bea
\benefit \ge
\lim_{\beta\to\infty}
\sum_{c=1}^{\Nclasses}
\left\langle
\left[\mathbb{P}_c(\wW_A\vec{X}) - \mathbb{P}_c(\beta\wWbar_A\vec{X})\right]
\Big\lbrace
[\wWbase_A\vec{X}]_c - [\wW_A\vec{X}]_c 
\Big\rbrace
\right\rangle 
\label{lowerboundMTL}
\eea

Similarly, taking expectations in (\ref{upperbound}) gives

\bea
\benefit \le
\lim_{\beta\to\infty}
\sum_{c=1}^{\Nclasses}
\left\langle
\left[\mathbb{P}_c(\wWbase_A\vec{X}) - \mathbb{P}_c(\beta\wWbar_A\vec{X})\right]
\Big\lbrace
[\wWbase_A\vec{X}]_c - [\wW_A\vec{X}]_c 
\Big\rbrace
\right\rangle
\label{upperboundMTL} 
\eea

Using Gaussian integration by parts in (\ref{lowerboundMTL}, \ref{upperboundMTL}), we obtain, after some straightforward algebra that

\bea
\benefit
\ge 
\Tr
\left(
\left[
{\G(\wWbar_A)}\wWbar_A - \G(\wW_A)\wW_A
\right]
\covX
\left[
\wW_A - \widetilde{\wW}_A
\right]^\dagger
\right)
\label{benefit_lower}
\eea

and 
\setlength{\abovedisplayskip}{-2pt}
\bea
\benefit
\le 
\Tr
\left(
\left[
{\G(\wWbar_A)}\wWbar_A - \G(\wWbase_A)\wWbase_A
\right]
\covX
\left[
\wW_A - \wWbase_A
\right]^\dagger 
\right)
\label{benefit_upper}
\eea

where $\G$ is defined above in (\ref{Gmatrix_defn}) via $\displaystyle{\left\langle X_k\mathbb{P}_c(\wW\vec{X})\right\rangle = \left[\G(\wW)\wW\covX\right]_{ck}}$\,.


These expressions are completely general and do not assume TA initialization or make any other approximations other than the assumptions stated above, ({\em viz.}~centered Gaussian random vectors with covariance $\covX$, linear activations in the hidden layers and a softmax in the output layer). 

It is also worth noting that the results hold for models of any depth since the $\wW$'s refer to the composite weight of the entire network.

Specializing to the TA case, following Appendix~\ref{appendix:TAmodel} above, and elaborated further in Appendix~\ref{appendix:TArank1benefit} gives the result quoted in our paper.

\subsection{Multitask Benefit for TA Networks} \label{appendix:TArank1benefit}
In order to address the multitask benefit for TA models, we need an extension of the single task analysis in Appendix~\ref{appendix:TAmodel} for multitask TA models. For simplicity, we consider the case where the data is drawn from 
a Gaussian distribution with $\covX = \boldsymbol{I}$.
 
 Recall that we defined the TA models by insisting that the student's initial weights have an SVD with the same singular vectors as in the teacher's SVD. The same definition applies here, so that 
 if $\noisyTeacher_{A/B} = \noisyU_{A/B} \noisyS_{A/B} \noisyVtt_{\hspace*{-2mm}AB}$ denotes the teachers' SVDs for tasks A and B, then the SVDs for the students' initial weights for tasks A and B are respectively set to

 $$
 \wW_A(0) = \noisyU_{A} \SwW_{A} \noisyVtt_{\hspace*{-2mm}A} \qquad \mbox{and} \qquad  \wW_B(0) = \noisyU_{B} \SwW_{B} \noisyVtt_{\hspace*{-2mm}B}
 $$

Using the definition of task relatedness, $\boldsymbol{r}_{AB} = \noisyVtt_{\hspace*{-2mm}B}\noisyV_{A} $, in the previous expression gives

 $$
  \wW_A(0) = \noisyU_{A} \SwW_{A} \noisyVtt_{\hspace*{-2mm}A} \qquad \mbox{and} \qquad 
 \wW_B(0) = \noisyU_{B} \SwW_{B}\boldsymbol{r}_{AB} \noisyVtt_{\hspace*{-2mm}A}.
 $$
 
As in Appendix~\ref{appendix:TAmodel}, singular vectors corresponding to the composite weight matrices can be written as the Hadamard product of the singular vectors corresponding to the layerwise weights. For example, for a model with a single hidden layer, we have $\mathrm{diag}(\SwW_A) := \ssw{A}\odot\ssW$ and $\mathrm{diag}(\SwW_B) := \ssw{B}\odot\ssW$. 
 
With these definitions, we can take the single task results for The TA model from Appendix~\ref{appendix:TAmodel} and extend them to two teachers to obtain

 \bea
 \nonumber 
\tau\frac{d}{dt}\ssw{A} 
&=& 
  {\boldsymbol{s}^{21}} \odot 
  \Big(  \boldsymbol{\hat{s}}_A g(\noisyTeacher_A\vert\noisyU_A)  -   
 \ssw{A}\odot\boldsymbol{s}^{21}  g(\wW_A\vert\noisyU_A)  \Big)
 \\ \nonumber
\tau\frac{d}{dt}\ssw{B}
 &=& 
  {\boldsymbol{s}^{21}} \odot 
   \Big(  \boldsymbol{\hat{s}}_Bg(\noisyTeacher_B\vert\noisyU_B)  -   
 \ssw{B}\odot\boldsymbol{s}^{21}  g(\wW_B\vert\noisyU_B)  \Big) \relatedness
\\ 
\boldsymbol{s}^{21} \odot \boldsymbol{s}^{21}
&=& 
\ssw{A} \odot \ssw{A} + \relatedness \ssw{B} \odot \ssw{B} 
 \label{multi_s_updates}
\eea
 
where $g(\wW\vert\boldsymbol{\hat{U}}) :=\transposed{\boldsymbol{\hat{U}}}G(\wW)\boldsymbol{\hat{U}}$ for any pair of matrices $\boldsymbol{\hat{U}}, \wW$.

Note that, as expected, if $\relatedness = 0$, the dynamics for the second task is trivial and only the first task evolves non-trivially. Thus, to obtain the single task dynamics, we can simply look at the case $\relatedness=0$. This motivates the following definition.

\begin{enumerate}
\item
Define $\bs(\relatedness)$ via the relation 

\be
\bs(\relatedness)\odot\bs(\relatedness) := \ssw{A}\odot \ssw{A}  \odot \left[ \ssw{A} \odot \ssw{A} + \relatedness \ssw{B} \odot \ssw{B} \right]
\label{srAB}
\ee

This simply says that $\lbrace {s_\sigma(\relatedness)} \rbrace$ are the multitask student's singular values pertinent to executing task A. 

\item
If we set $\relatedness=0$ in (\ref{srAB}), we recover the dynamics of the single task case. Therefore, if 
$\lbrace \tilde{s}_\sigma \rbrace$ denote the student's singular values when training on task A, we can identify 

$$
\widetilde{\bs} = \bs(\relatedness=0)
$$
\end{enumerate}

To understand the difference between the multitask and single task case, we need to consider how $\relatedness$ modifies the results in the single task case. The simplest way to do this is to study equation~(\ref{multi_s_updates}) perturbatively in 
$\relatedness$ by plugging in the last line of (\ref{multi_s_updates}) into the first line of (\ref{multi_s_updates}). The mechanics of carrying out the perturbative expansion, while somewhat tedious, are straightforward and are left as an exercise for the motivated reader. The result of the exercise can be summarized as follows:

\begin{enumerate}
\item
Let $\lbrace \hat{s}^{A}_\sigma \rbrace$ denote the singular values of the teacher corresponding to task A.  
By definition, the initial singular values corresponding to the multitask student's parameters for task A and those of the baseline single task student are identical. With these initial conditions, SGD dynamics yields, at all times,  

\bea
s_\sigma(\relatedness) &\ge& s_\sigma(\relatedness=0) = \tilde{s}_\sigma \qquad \mbox{whenever ${\tilde{s}_\sigma}{\vert_{t=0}} < \hat{s}^{A}_\sigma$}\nonumber\\
s_\sigma(\relatedness) &\le& s_\sigma(\relatedness=0) = \tilde{s}_\sigma \qquad \mbox{whenever ${\tilde{s}_\sigma}{\vert_{t=0}} \ge \hat{s}^{A}_\sigma$}\nonumber\\
\label{singularvalsdynamics}
\eea

\vst
In other words, the effect of $\relatedness>0$ is to enhance either the growth rate or the decay rate of the student's singular values along the ``principle components'' of the noisy teacher.

\item
Let $\lbrace \sbarA_\sigma\rbrace$ denote the singular values of the noise-free teacher.

\begin{enumerate}
\item
High SNR Case:

When the SNR for task A is large, the singular vectors of the noise-free teacher are almost surely parallel to the singular values of the noisy teacher (see \cite{BenaychGeorges}). In this case, 

\bea
\Tr
\left(
\left[
{\G(\wWbar_A)}\wWbar_A
\right]
\left[
\wW_A - \widetilde{\wW}_A
\right]^\dagger
\right)
= 
\sum_{\sigma=1}^{\mbox{ {\footnotesize rank$_A$}}}
g(\overline{\bs}_A) \sbarA_\sigma 
\big( s_\sigma(\relatedness)-\tilde{s}_\sigma\big) \ge 0
\label{noisefree}
\eea

where $\rank_A:=\rank(\wWbar_A)$ is the rank of the noise-free teacher. Consequently, equation (\ref{benefit_lower}) yields

\bea
\benefit
&\ge&
\sum_{\sigma=1}^{\mbox{ {\footnotesize rank$_A$}}}
\left[
g(\overline{\bs}_A) \sbarA_\sigma 
-
g(\bs(\relatedness)) s_\sigma(\relatedness) 
\right]
\big[ s_\sigma(\relatedness)-\tilde{s}_\sigma\big]
\nonumber\\
&&
\,\,\, + 
\sum_{\sigma > \rank_A}
g(\bs(\relatedness)) s_\sigma(\relatedness) 
\big\vert s_\sigma(\relatedness)-\tilde{s}_\sigma\big\vert
\nonumber\\
\label{benefit_hiSNR}
\eea

Thus, from the assertion in (\ref{dgdS}), 

\be
\Vert\bs(\relatedness)\Vert > \Vert\overline{\bs}_A\Vert 
\qquad \Rightarrow \qquad
\benefit > 0
\label{hiSNR}
\ee

\item
Low SNR Case:

When the SNR for task A is small, the singular vectors of the noise-free teacher are almost surely orthogonal to the singular values of the noisy teacher (see \cite{BenaychGeorges}). In this case, 

\bea
\Tr
\left(
\left[
{\G(\wWbar_A)}\wWbar_A
\right]
\left[
\wW_A - \widetilde{\wW}_A
\right]^\dagger
\right)
= 0
\eea

Consequently, equations (\ref{benefit_lower}, \ref{benefit_upper}) yield

\bea
\sum_{\sigma=1}^{\Nclasses}
g(\bs(\relatedness)) s_\sigma(\relatedness) 
\big[\tilde{s}_\sigma- s_\sigma(\relatedness)\big]
\le
\benefit
\le
\sum_{\sigma=1}^{\Nclasses}
g(\widetilde{\bs}) \tilde{s}_\sigma
\big[\tilde{s}_\sigma- s_\sigma(\relatedness)\big]
\nonumber\\
\label{benefit_lowSNR}
\eea

\par\vst
On the other hand,  in the low SNR case, the singular values of the noisy teacher are essentially in the bulk of the MP sea (cf. \cite{BenaychGeorges}). Therefore, according to equation (\ref{singularvalsdynamics}),  the sign of the multitask benefit in the low SNR regime will depend on the size of the set  $\big\lbrace \sigma \big\vert \tilde{s}_\sigma{\vert_{t=0}} \ge {\hat{s}^{A}}_\sigma\big\rbrace$ where the ${\hat{s}^{A}}_\sigma$ are drawn from the MP distribution. When this set is small, 
$\benefit$ will tend to be negative. Conversely $\benefit$ will tend to be positive if the size of the forementioned set is large. 

\end{enumerate}

\item
Note that according equation (\ref{multi_s_updates}), if $\relatedness>0$, then 
\begin{enumerate}
    \item increasing $\overline{\bs}_B$ (the SNR of task B) increases $\hat{\bs}_B$.
    \item According to (\ref{dgdS}), increasing $\Vert\hat{\bs}_B\Vert$ decreases $g(\hat{\bs}_B)\hat{\bs}_B$, 
    which in turn decreases the rate of growth of $\bs(\relatedness)$ relative to the rate of growth of 
$\widetilde{\bs}$ (cf.~the second line in equation~\ref{multi_s_updates}).
\end{enumerate}
Hence, an increase in $\hat{\bs}_B$ results in an overall increase of $\tilde{s}_\sigma - s_\sigma(\relatedness)$ 
and consequently an increase in $\benefit$ in both the low SNR task A case following equation (\ref{benefit_lowSNR}) and the high SNR task A case where the second term in (\ref{benefit_hiSNR}) increases with  $\big\vert\tilde{s}_\sigma - s_\sigma(\relatedness)\big\vert$. In other words 

\be
\mbox{increasing  $\hat{\bs}_B\Big\vert_{{\tiny \relatedness >0}}$} \Rightarrow \mbox{an increase in  $\big[\tilde{s}_\sigma - s_\sigma(\relatedness)\big]$ } \Rightarrow \mbox{an increase in $\benefit$}
\label{benefit_vs_sB}
\ee

Note that for small but nonzero values of $\relatedness$ and very large values of $\Vert\hat{\bs}_B\Vert$, $g(\hat{\bs}_B)\hat{\bs}_B \to 0$ so that the second line of equation~(\ref{multi_s_updates}) leads to a rapid decay of $\ssw{B}$ towards zero, which in turn implies that $\bs(\relatedness)- \widetilde{\bs}$ becomes negative following the third line of equation~(\ref{multi_s_updates}). Thus, 
$\big[\tilde{s}_\sigma - s_\sigma(\relatedness)\big] \to c_\sigma \le 0$ for $\Vert\hat{\bs}_B\Vert \gg 1$. Consequently, regardless of the SNR for task A, 

\be
\mbox{$\benefit \to m \le 0$ for very large values of $\Vert\hat{\bs}_B\Vert$ and small $\relatedness>0$}.
\label{benefit_vs_very_large_sB}
\ee

\item
We could also increase $\relatedness$ with $\overline{\bs}_B>0$ fixed. 

The third line of equation~(\ref{multi_s_updates}) shows that $\bs(\relatedness)$ monotonically increases with $\relatedness>0$. This in turn directly implies that the differences in the components of 
$[\bs(\relatedness)-\widetilde{\bs}]$ will all increase as $\relatedness$ increases. Therefore, in the high SNR regime for task A, equation~(\ref{benefit_hiSNR}) immediately gives 

\be
\mbox{an increase $\relatedness$} \Rightarrow \mbox{an increase in $\benefit$}
\label{benefit_vs_rAB}
\ee

\item
Finally, we note that as $\Ndata \to \infty$, the empirical cross-covariance between the noisy labels and the input feature vectors converges to true cross-covariance between the noise-free labels and the input feature vectors as long as the 
noise level remains bounded by $\Nclasses/\Nf$. In this case, the generalization loss and the training loss are almost surely equal so that 

\be
\lim_{\Ndata\to\infty}\mathcal{L}_{A|B} \to \mathcal{L}_{A} \Rightarrow \lim_{\Ndata\to\infty} \benefit \to 0.
\label{infinite_data}
\ee

\end{enumerate}

\clearpage
\section{Task Relatedness and Multitask Results}
\setlength{\abovedisplayskip}{-6pt}

\subsection{Multitask full results}
\begin{figure}[H]
    \centering
    \includegraphics[]{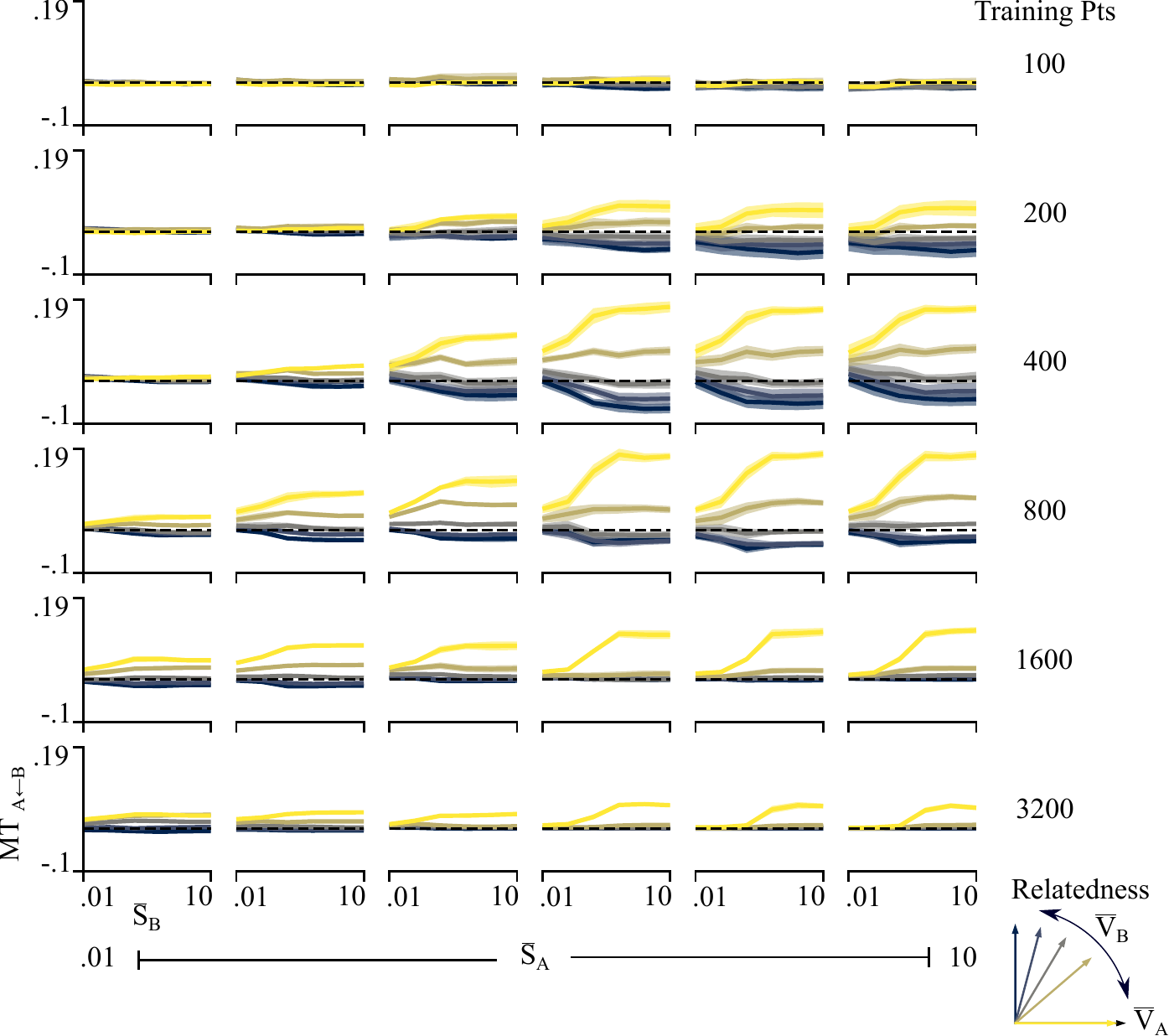}
    \caption{Multitask benefit in student networks trained on data that varied along 4 independent variables: 1) number of main task training points (rows), 2) main task signal-to-noise ratio (SNR) (columns), 3) auxiliary task SNR (x-axes), and 4) auxiliary task relatedeness (individual lines). Each line shows the mean benefit over 5 random seeds and the shaded region shows the standard error of the mean. Multitask benefits > 0 indicate that student network performs better when trained with additional auxiliary task data. MT benefit is correlated with task relatedness and SNR for related tasks, yet negatively correlated with SNR for unrelated tasks. This data is summarized in Figure \ref{fig:multi_main}.}
    \label{appendix:rank10}
\end{figure}
\begin{figure}[H]
  \centering
  \includegraphics[]{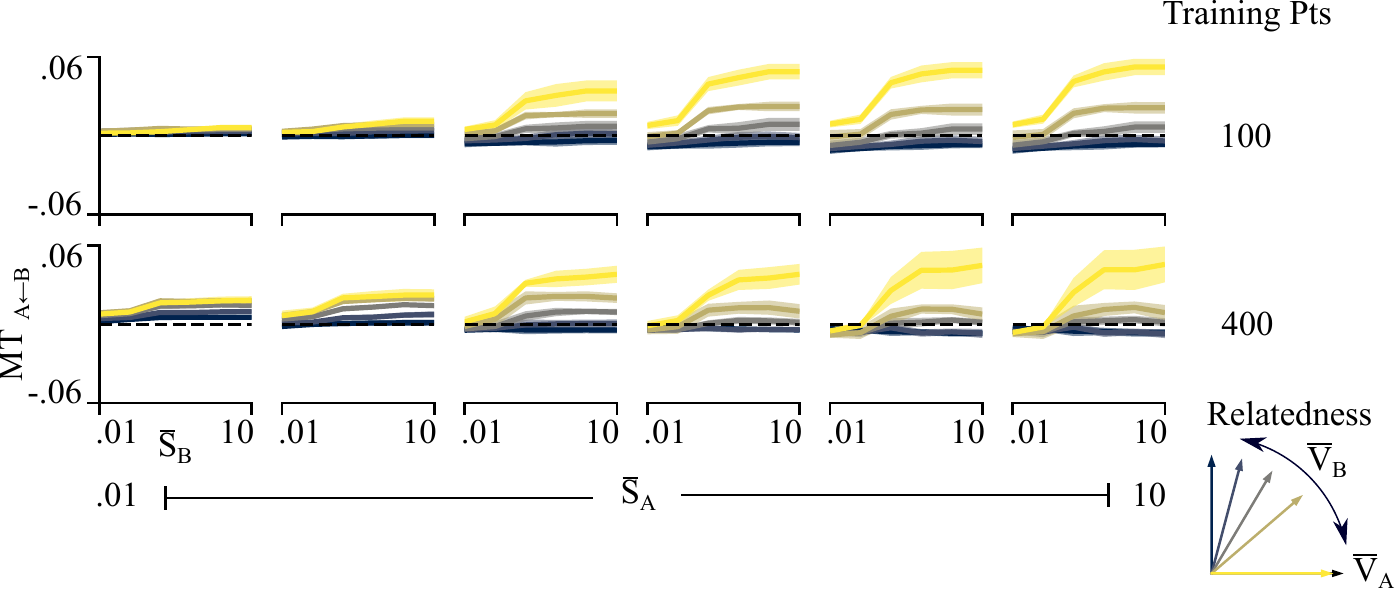}
  \caption{Multitask benefit when trained on rank 3 teachers. The data is arranged as in Figure \ref{appendix:rank10} and shows very similar trends as in the rank 10 case.}
  \label{appendix:rank3}
\end{figure}
\begin{figure}[H]
  \centering
  \includegraphics[]{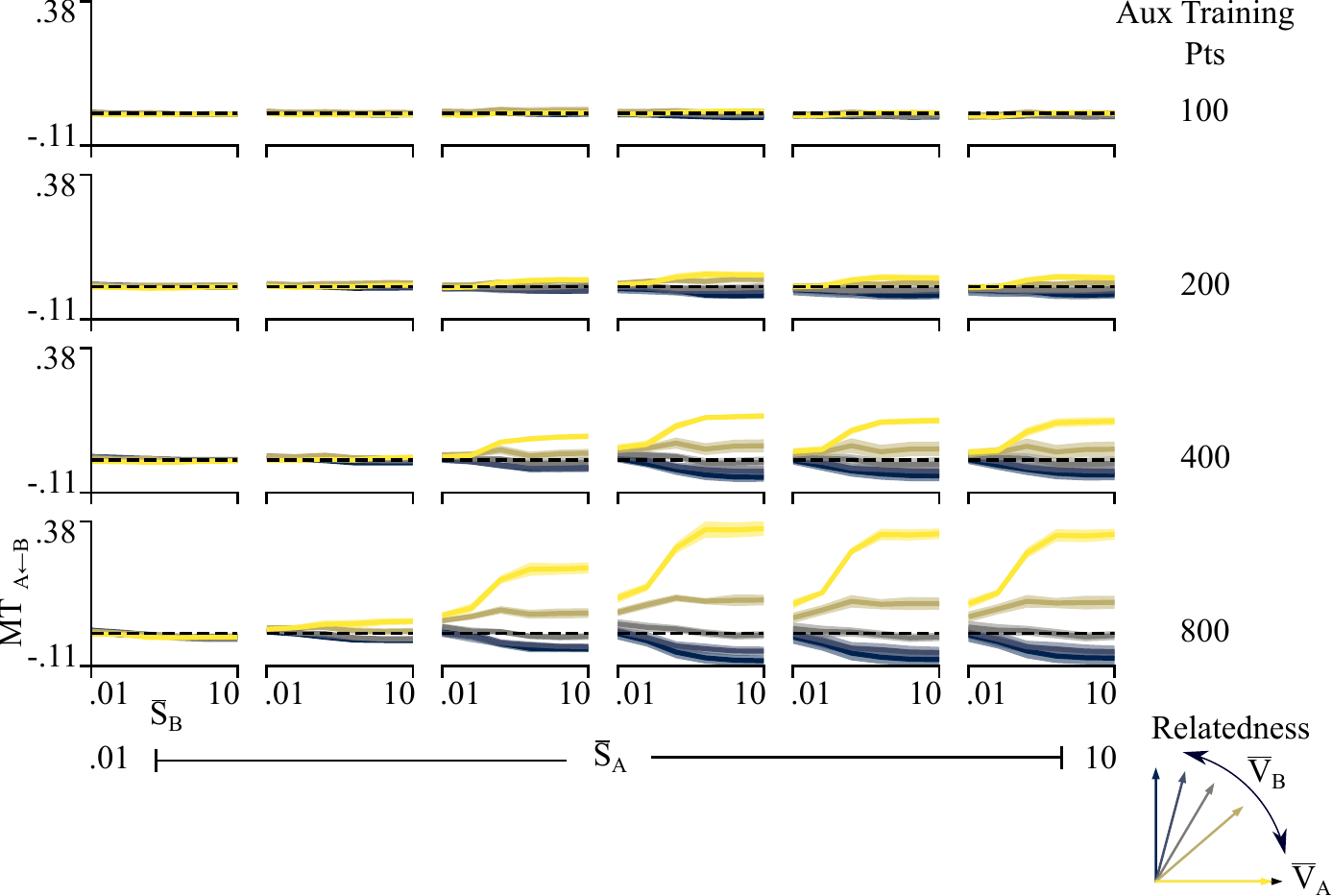}
  \caption{Multitask benefit when trained on increasing levels of auxiliary task data. The data is arranged as in Figure \ref{appendix:rank10} and shows greatly improved performance with larger amounts of auxiliary data. This data is summarized in Figure \ref{fig:multi_aux_data}, left panel.}
  \label{appendix:multi_aux_data}
\end{figure}
\begin{figure}[H]
  \centering
  \includegraphics[]{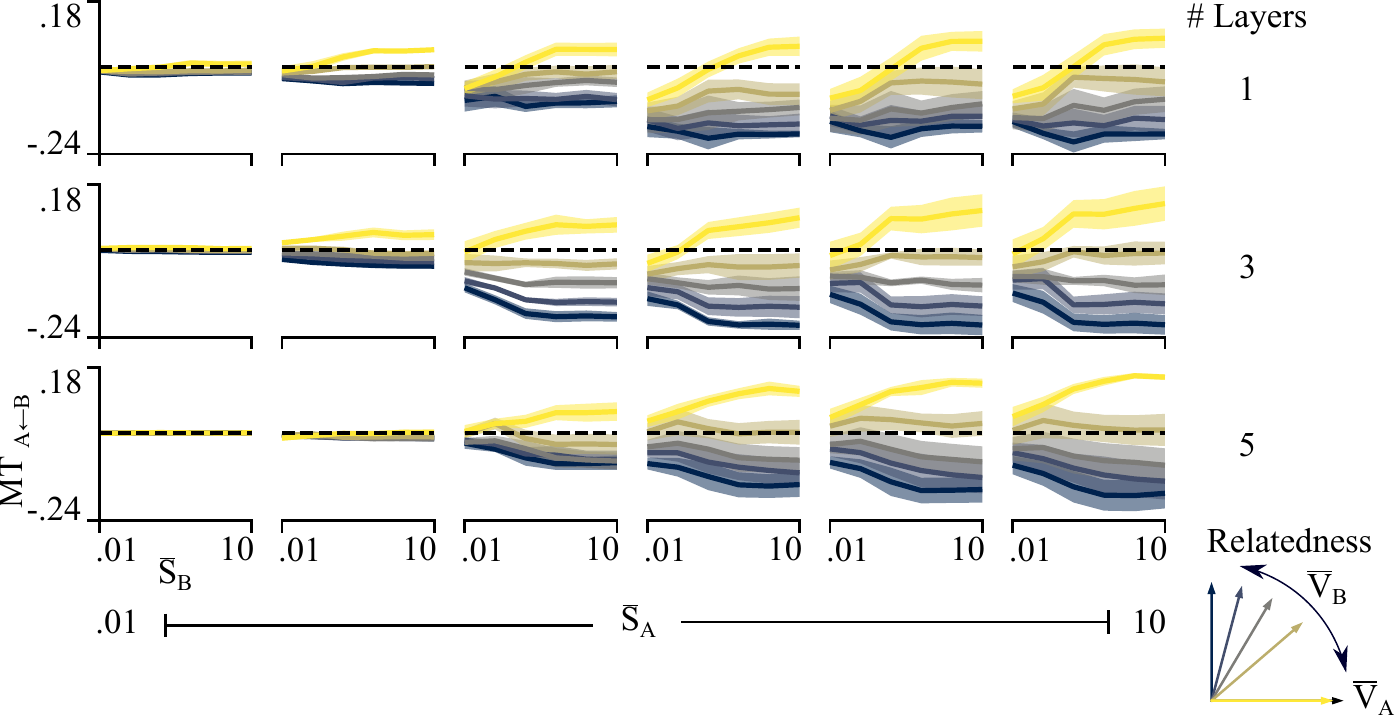}
  \caption{Multitask benefit when training deeper networks with ReLU nonlinearities. The data is arranged as in Figure \ref{appendix:rank10} and shows qualitatively similar results to linear networks. This data is summarized in Figure \ref{fig:multi_aux_data}, right panel.}
  \label{appendix:multi_nonlinear}
\end{figure}

\clearpage
\end{appendices}

\end{document}